\documentclass[sigconf]{lib/acmart}

\usepackage{array}
\usepackage{latexsym}
\usepackage{amsmath}
\usepackage{amsthm}
\usepackage{bm}
\usepackage{xspace}
\usepackage{amsfonts}
\usepackage{physics}
\usepackage{xcolor}
\usepackage{colortbl}
\usepackage{algorithm}
\usepackage{algorithmic}
\usepackage{url}
\usepackage{float}
\usepackage{booktabs}
\usepackage{multirow}
\usepackage{pifont}
\usepackage{siunitx}
\usepackage[subrefformat=parens]{subcaption}
\usepackage{xifthen}  
\usepackage{xparse}
\usepackage{lib/eclbkbox}

\newcommand{\TSK}[1]{#1}

\graphicspath{{./fig/}}

\newcommand{\bit}{\vspace{-0em}\begin{itemize}}
\newcommand{\eit}{\end{itemize}\vspace{-0.2em}}
\newcommand{\ben}{\vspace{-0em}\begin{enumerate}}
\newcommand{\een}{\end{enumerate}\vspace{-0.2em}}
\newcommand{\bea}{\vspace{-0em}\begin{eqnarray}}
\newcommand{\eea}{\end{eqnarray}\vspace{-0.0em}}
\newcommand{\beq}{\vspace{-0.0em}\begin{equation}}
\newcommand{\eeq}{\end{equation}\vspace{-0.0em}}

\newcommand{\argmin}{\mathop{\rm arg~min}}

\newcommand{\R}{\mathbb{R}}
\newcommand{\C}{\mathbb{C}}

\newcommand{\hide}[1]{}
\newcommand{\myparaitemize}[1]{\noindent{\textbf{#1.}}}

\newtheorem{problem}{Problem}
\newtheorem{lemma}{\textsc{Lemma}}

\newtheorem{definition}{Definition}
\newtheorem{model}{Model}

\newcommand{\relation}{time-evolving causality\xspace}

\newcommand{\Relation}{Time-evolving causality\xspace}
\newcommand{\RELATION}{Time-evolving Causality\xspace}

\newcommand{\mat}[1]{\bm{#1}}

\newcommand{\mB}{\mat{B}}
\newcommand{\mE}{\mat{E}}
\newcommand{\mH}{\mat{H}}
\newcommand{\mX}{\mat{X}}
\newcommand{\mW}{\mat{W}}

\newcommand{\vect}[1]{\bm{#1}}
\newcommand{\rowvect}[1]{\vect{#1}}

\newcommand{\vx}{\vect{x}}

\newcommand{\vs}{\vect{s}}
\newcommand{\vvec}{\vect{v}}

\newcommand{\latent}{\vect{s}}
\newcommand{\ind}{\vect{e}}

\newcommand{\est}{\vect{v}}
\newcommand{\demixing}{\mat{W}}

\newcommand{\regime}{\boldsymbol{\theta}}
\newcommand{\regimeset}{\boldsymbol{\Theta}}
\newcommand{\update}{\boldsymbol{\omega}}
\newcommand{\updateset}{\boldsymbol{\Omega}}
\newcommand{\modelparam}{\mathcal{F}}
\newcommand{\candparam}{\mathcal{C}}
\newcommand{\selfdynamics}{\mathcal{D}}

\newcommand{\hankel}{\mat{H}}
\newcommand{\embed}[1]{{\textit{g}({#1})}}
\newcommand{\invembed}[1]{{\textit{g}^{-1}({#1})}}

\newcommand{\nmodes}{k}
\newcommand{\modes}{\mat{\Phi}}
\newcommand{\eigs}{\mat{\Lambda}}

\newcommand{\imode}{\ith{\mat{\Phi}}}
\newcommand{\ieig}{\ith{\mat{\Lambda}}}

\newcommand{\trans}{\mat{A}}

\newcommand{\forgetting}{\mu}
\newcommand{\Forgetting}{\mat{M}}
\newcommand{\ith}[2][]{%
    \ifthenelse{\isempty{#1}}%
    {
        #2_{(i)}
    }%
    {
        {#2_{(i)}^{#1}}
    }%
}
\newcommand{\first}[2][]{%
    \ifthenelse{\isempty{#1}}%
    {
        #2_{(1)}
    }%
    {
        {#2_{(1)}^{#1}}
    }%
}
\newcommand{\dth}[2][]{%
    \ifthenelse{\isempty{#1}}%
    {
        #2_{(d)}
    }%
    {
        {#2_{(d)}^{#1}}
    }%
}

\newcommand{\method}{\textsc{ModePlait}\xspace}
\newcommand{\modelestimator}{\textsc{ModeEstimator}\xspace}
\newcommand{\modelgenerator}{\textsc{ModeGenerator}\xspace}

\newcommand{\regimeupdate}{\textsc{RegimeUpdater}\xspace}

\newcommand{\synthetic}{\textit{\#0 synthetic}\xspace}
\newcommand{\psynthetic}{\textit{(\#0) synthetic}\xspace}
\newcommand{\covid}{\textit{\#1 covid19}\xspace}
\newcommand{\pcovid}{\textit{(\#1) covid19}\xspace}
\newcommand{\googletrend}{\textit{\#2 web-search}\xspace}
\newcommand{\pgoogletrend}{\textit{(\#2) web-search}\xspace}
\newcommand{\chickendance}{\textit{\#3 chicken-dance}\xspace}
\newcommand{\pchickendance}{\textit{(\#3) chicken-dance}\xspace}
\newcommand{\exercise}{\textit{\#4 exercise}\xspace}
\newcommand{\pexercise}{\textit{(\#4) exercise}\xspace}

\AtBeginDocument{%
  \providecommand\BibTeX{{%
    \normalfont B\kern-0.5em{\scshape i\kern-0.25em b}\kern-0.8em\TeX}}}

\copyrightyear{2025}
\acmYear{2025}
\setcopyright{acmlicensed}
\acmConference[KDD '25]{Proceedings of the 31st ACM SIGKDD Conference on Knowledge Discovery and Data Mining V.1}{August 3--7, 2025}{Toronto, ON, Canada}
\acmBooktitle{Proceedings of the 31st ACM SIGKDD Conference on Knowledge Discovery and Data Mining V.1 (KDD '25), August 3--7, 2025, Toronto, ON, Canada}
\acmDOI{10.1145/3690624.3709283}
\acmISBN{979-8-4007-1245-6/25/08}

\begin{document}

\title{Modeling Time-evolving Causality over Data Streams}

\author{Naoki Chihara}
\affiliation{%
  \institution{SANKEN, Osaka University}
  \state{Osaka}
  \country{Japan}
}
\email{naoki88@sanken.osaka-u.ac.jp}

\author{Yasuko Matsubara}
\affiliation{%
  \institution{SANKEN, Osaka University}
  \state{Osaka}
  \country{Japan}
}
\email{yasuko@sanken.osaka-u.ac.jp}

\author{Ren Fujiwara}
\affiliation{%
  \institution{SANKEN, Osaka University}
  \state{Osaka}
  \country{Japan}
}
\email{r-fujiwr88@sanken.osaka-u.ac.jp}

\author{Yasushi Sakurai}
\affiliation{%
  \institution{SANKEN, Osaka University}
  \state{Osaka}
  \country{Japan}
}
\email{yasushi@sanken.osaka-u.ac.jp}


\begin{abstract}
Given an extensive,
semi-infinite collection of multivariate co-evolving data sequences
(e.g., sensor/web activity streams)
whose observations influence each other,
how can we discover the time-changing cause-and-effect relationships in co-evolving data streams?
How efficiently can we reveal dynamical patterns that allow us to forecast future values?
In this paper,
we present a novel streaming method, \method,
which is designed for modeling such causal relationships (i.e., \relation)
in multivariate co-evolving data streams
and forecasting their future values.
The solution relies on characteristics of the causal relationships that evolve over time in accordance with the dynamic changes of exogenous variables.
\method has the following properties: 
(a) \textit{Effective}:
it discovers the \relation in multivariate co-evolving data streams by detecting the transitions of distinct dynamical patterns adaptively.
(b) \textit{Accurate}: it enables both the discovery of \relation and the forecasting of future values in a streaming fashion.
(c) \textit{Scalable}:
our algorithm does not depend on data stream length
and thus is applicable to very large sequences.
Extensive experiments on both synthetic and real-world datasets demonstrate that our proposed model outperforms state-of-the-art methods in terms of discovering the \relation as well as forecasting.

\end{abstract}




\maketitle

\section{Introduction}
    \label{section:introduction}
    In recent years,
a substantial quantity of multivariate time series data
has been collected from various events and applications
related to the Internet of Things (IoT)~\cite{de2016iot,mahdavinejad2018machine},
web activities~\cite{kawabata2020non,nakamura2023fast},
the spread of infectious diseases~\cite{kimura2022fast}, and
patterns of user behavior~\cite{matsubara2012fast}.
In real-world scenarios, in particular,
these data are generated quickly and continuously, making it increasingly important to process them in a streaming fashion.
\par
There are various relationships between observations
in time series (e.g., correlation, independency)
and they are critical features for a wide range of time series analyses,
including clustering~\cite{hallac2017toeplitz, obata2024dynamic}, forecasting~\cite{Yuqietal-2023-PatchTST}, imputation~\cite{wang2023networked}, and others~\cite{tozzo2021statistical}.
Among them, causality
\cite{bollen1989structural, spirtes2000causation}
offers particularly valuable insight,
with many studies dedicated to its investigation
\cite{he2021daring, fujiwara2023causal}
and improving downstream tasks by employing it as an inductive bias, guiding the learning process towards more generalizable representations
\cite{dai2022graph, cheng2023cuts}.
However, most existing methods assume that causal relationships do not evolve over time in multivariate time series~\cite{runge2018causal}.
It is crucial to discover such causal relationships
if we are to detect new causative factors promptly
and accurately forecast in a streaming fashion.
Their pivotal role becomes increasingly apparent upon
recognizing that real-world data streams contain these relationships.
For example,
with the spread of infectious diseases,
when a new virus strain emerges in a country,
certain activities, such as cross-border travel,
can lead to an increase in the number of infections in other countries,
and the causative countries change over time.
Here, we refer to such time-changing causal relationships as ``\textit{\relation}.''
So, how can we discover the \relation in data streams and 
model semi-infinite multivariate data sequences?
\par
Briefly, we design our model based on the structural equation model (SEM)~\cite{pearl2009causality}, 
which is often utilized for learning causal structures.
This model consists of two types of variables:
endogenous variables and exogenous variables.
The former are determined by the model itself,
while the latter exist outside the model.
In addition,
exogenous variables are independent of each other
and follow a non-Gaussian distribution.
In our work, we consider that the causal relationships in data streams evolve over time in accordance with the dynamic changes of exogenous variables.
\par
There are two difficulties involved in
designing a model for discovering \relation.
\textit{(i) Latent temporal dynamics in univariate time series}:
As pointed out above,
exogenous variables in the structural equation model
are independent of each other,
so it is unsuitable to consider
them as multivariate time series.
Therefore, we need to capture latent temporal dynamics in univariate time series
to discover the \relation.
However, it is challenging to design an appropriate system for univariate time series
because the latent temporal dynamics in the system are generally multi-dimensional,
and so a single dimension is insufficient for modeling the system.
We solve this issue by expressing univariate time series
as the superposition of computed basis vectors (i.e., modes) associated with
decay rate and temporal frequency.
\textit{(ii) Distinct dynamical patterns}:
Data streams typically contain various types of distinct dynamical patterns, and
they are factors that change causal relationships over time.
It is essential to understand their changes
if we are to model an entire data stream effectively.
For example,
in the context of web search activities,
we can identify various types of pattern changes
caused by a multitude of reasons, such as a new item release.
In addition, the event influences sales of other items, so the causal relationships could also change.
We refer to these distinct dynamical patterns in data streams as ``\textit{regimes}.''
\par
\TSK{
\begin{figure}
    \begin{tabular}{cc}
      \label{fig:crown:causal}
      \hspace{-0.5em}
      \begin{minipage}[c]{0.47\linewidth}
        \centering
        \includegraphics[width=0.78\linewidth]{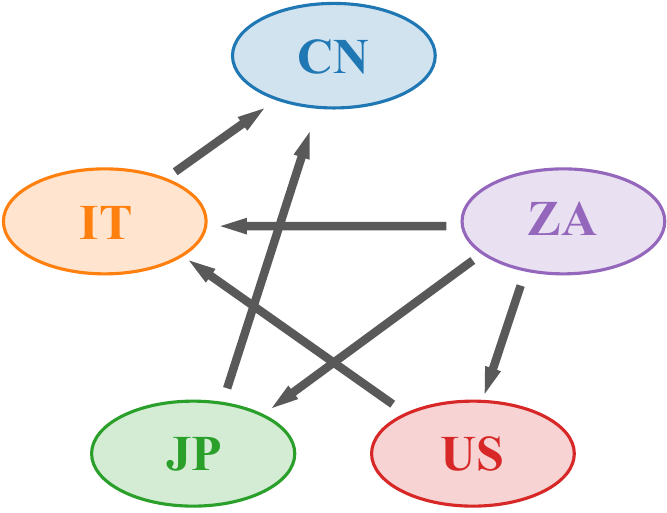}
        \vspace{0.1em}
        \\
        (a-i) January 8, 2021
      \end{minipage} &
      \hspace{-1.5em}
      \begin{minipage}[c]{0.47\linewidth}
        \centering
        \includegraphics[width=0.78\linewidth]{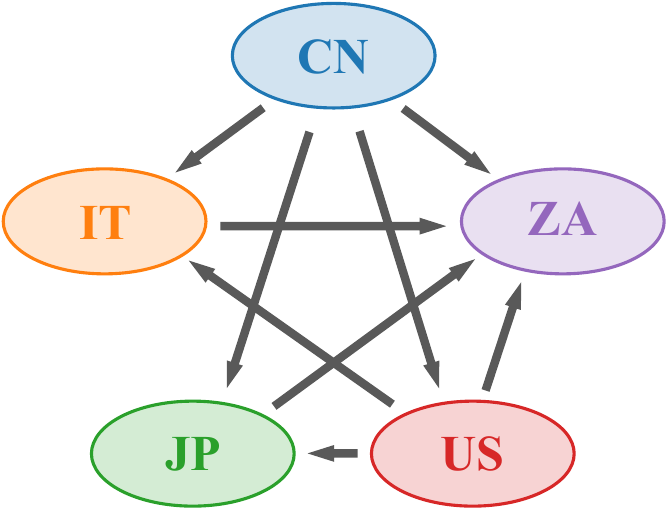}
        \vspace{0.1em}
        \\
        (a-ii) May 19, 2022
      \end{minipage}
      \vspace{0.4em}
      \\
      \multicolumn{2}{c}{\textbf{(a) Causal relationships at different time points}}
      \\
      \vspace{-0.8em}
      \end{tabular}
      \begin{tabular}{ccc}
        \label{fig:crown:latent}
        \hspace{-0.5em}
        \begin{minipage}[c]{0.32\linewidth}
          \centering
          \includegraphics[width=0.65\linewidth]{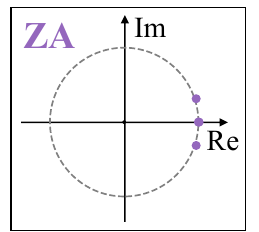}
          \\
          (b-i) Dec. 6, 2020
        \end{minipage} &
        \hspace{-1.5em}
        \begin{minipage}[c]{0.32\linewidth}
          \centering
          \includegraphics[width=0.65\linewidth]{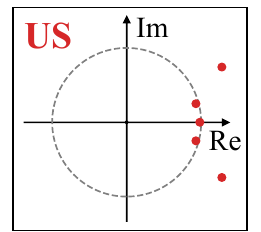}
          \\
          (b-ii) Jan. 10, 2022
        \end{minipage} &
        \hspace{-1.5em}
        \begin{minipage}[c]{0.32\linewidth}
          \centering
          \includegraphics[width=0.65\linewidth]{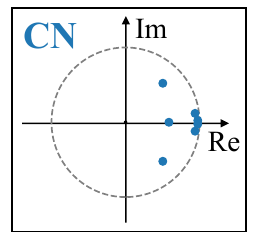}
          \\
          \vspace{-0.08em}
          (b-iii) May 19, 2022
        \end{minipage} \vspace{0.4em} \\
      \multicolumn{3}{c}{\textbf{(b) Eigenvalues of latent temporal dynamics}}
      \vspace{0.35em}
    \end{tabular}
    \begin{tabular}{cc}
      \label{fig:crown:causal}
      \hspace{-1.7em}
      \begin{minipage}[c]{0.49\linewidth}
        \centering
        \includegraphics[width=0.91\linewidth]{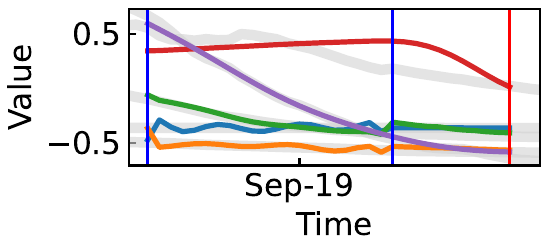}
        \vspace{-0.3em}
        \\
        \hspace{2.0em}
        (c-i) September 27, 2021
      \end{minipage} &
      \hspace{-1.7em}
      \begin{minipage}[c]{0.49\linewidth}
        \centering
        \includegraphics[width=0.88\linewidth]{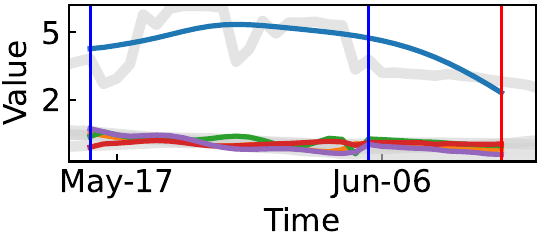}
        \vspace{-0.35em}
        \\
        (c-ii) June 5, 2022
      \end{minipage}
      \vspace{0.4em}
      \\
      \multicolumn{2}{c}{\textbf{(c) Snapshots of 10 days-ahead future value forecasting}}
      \\
      \vspace{-1.2em}
      \end{tabular}
    \vspace{-1.0em}
    \caption{Modeling power of \method over an epidemiological data stream (i.e., \covid):
    This original stream consists of daily COVID-19 infection numbers in five major countries.
    Our proposed method can (a) discover the causal relationships, which change over time, (b) extract the eigenvalues of the latent dynamics providing insight into them in terms of decay rate and temporal frequency, and (c) forecast future value in a stream fashion.}
    \label{fig:crown}
    \vspace{1.0ex}
\end{figure}

}
In this paper, we present \method\footnote{Our source codes and datasets are available at~\cite{code}},
which
simultaneously and continuously
discovers \relation in a multivariate co-evolving data stream
and forecasts future values.
In addition, 
thanks to desirable features of modes,
\method extracts
the temporal behavior of the latent dynamics in each exogenous variable in terms of decay rate and temporal frequency by analyzing the corresponding eigenvalues.
\\ In short, the problem we deal with is as follows:
\vspace{1.15mm} \\
\textit{\noindent\textbf{Given:} a semi-infinite multivariate data stream $\mat{X}$, which consists of $d$-dimensional vectors $\vect{x}(t)$, i.e., $\mat{X} = \{ \vect{x}(1), ..., \vect{x}(t_c), ... \}$, where $t_c$ is the current time,
{\setlength{\leftmargini}{24pt}
\begin{itemize}
\item \textbf{Find} distinct dynamical patterns (\textit{i.e., regimes}),
\item \textbf{Discover} causal relationships that changes in accordance with the transitions of regimes (\textit{i.e., \relation}),
\item \textbf{Forecast} an $l_s$-steps-ahead future value, i.e., $\vvec(t_c+l_s)$,
\end{itemize}
}
continuously and quickly, in a streaming fashion.
}\par\noindent
\vspace{-2.0em}
\subsection{Preview of Our Results}
Figure \ref{fig:crown} shows the results obtained with \method
for modeling an epidemiological data stream (i.e., \covid).
This dataset consists of the number of COVID-19 infected patients in five countries (i.e., Japan, the United States, China, Italy, and the Republic of South Africa).
Our method captures the following properties: \par
\par
\myparaitemize{\Relation}
Figure \ref{fig:crown} (a) shows
the causal relationships
between observations, which change over time.
The arrows indicate causality:
the base of each arrow represents the ``cause,''
while the head represents the ``effect.''
\method successfully discovers the time-changing causal relationships between countries from an epidemiological data stream.
For example,
Figure \ref{fig:crown} (a-i) shows that the Republic of South Africa had a causal influence on other countries.
This finding corresponds to the fact that
health officials announced the discovery of a new lineage of the coronavirus, namely 501.V2, in South Africa on December 18, 2020~\cite{covid19_africa},
and indicates that \method adaptively discovered the influence of the new coronavirus on other countries.
Additionally, Figure \ref{fig:crown} (a-ii) shows that China had a causal influence on other countries in contrast to Figure \ref{fig:crown} (a-i).
This insight aligns with the period of one of the longest and toughest lockdowns in Shanghai, which lasted from early April 2022 to June 1, 2022~\cite{covid19_china}.
It implies that
\method detected the influence of the spread of coronavirus infection in Shanghai,
which led to a strict and long-term lockdown.
In summary, the above discussions make it clear that
\method can capture the transitions of distinct dynamical patterns in an epidemiological data stream and
adaptively discover causal relationships at any given moment.
\par
\myparaitemize{Latent temporal dynamics}
Figure \ref{fig:crown} (b) shows the eigenvalues of latent temporal dynamics in exogenous variables.
These figures represent complex planes.
The dotted gray lines are unit circles, and
colored points are eigenvalues of latent temporal dynamics,
whose magnitude and argument indicate the decay rate and temporal frequency of a specific mode, respectively.
Specifically, if the absolute value of an eigenvalue is greater than $1$, the corresponding mode exhibits growth; if it is less than $1$, it exhibits decay
(please see Section \ref{section:uni} for a detailed approach to reading these components).
Figure \ref{fig:crown} (b-i) shows the weak growth modes in exogenous variables for the Republic of South Africa,
and implies an increase in infections in South Africa due to 501.V2 mentioned above.
Figure \ref{fig:crown} (b-ii) shows the strong growth modes in exogenous variables for the United States,
where new infections surpassed 1 million in a single day for the first time~\cite{covid19_america}.
Figure \ref{fig:crown} (b-iii) shows the decay of exogenous variables for China.
This period was toward the end of the lockdown in Shanghai, indicating that the spread of infections was beginning to ease,
our result captures this precisely.
\par
\myparaitemize{$l_s$-steps-ahead future values}
Figure \ref{fig:crown} (c)
shows snapshots of
the $l_s = 10$-steps-ahead future forecasting
when given a current window.
The blue vertical axes show the beginning of a current window $t_m$ and
the current time point $t_c$,
and the red vertical axis shows the $l_s$-step-ahead time point $t_c+l_s$.
In addition, we show our estimated values with bold-colored lines
(the originals are shown in gray).
\method successfully finds the current distinct dynamical pattern and
generates future values continuously at any time.
\subsection{Contributions} 
In this paper, we propose \method, which has all of the following desirable properties: \par
{\setlength{\leftmargini}{12pt}
\begin{itemize}
\item \textit{Effective}:
it discovers time-changing relationships between observations (i.e., \relation) based on monitoring transitions of distinct dynamical patterns (i.e., regimes).
\item\textit{Accurate}:
it theoretically discovers the \relation in data streams (please see Lemma \ref{lemma:causal} for details), and accurately forecast future values based on these relationships.
\item \textit{Scalable}:
it is fast and requires only constant computational time with regard to the entire stream length.
\end{itemize}
}


\section{Related Work}
    \label{section:related_work}
    \TSK{
\newcommand*\rot{\rotatebox{90}}

\newcommand*\OK{\ding{51}}
\newcommand*\NO{-}
\newcommand{\SOME}{some}

\begin{table}[t]
\vspace{-0.5em}
\centering
\caption{
Capabilities of approaches.
}
\label{table:capability}
\vspace{-1.2em} 
\scalebox{0.99}{
    \begin{tabular}{|l|ccccc|c|}
    \hline
    &
    \rot{ARIMA/++} &
    \rot{TICC} &
    \rot{NOTEARS/++} &
    \rot{OrbitMap} &
    \rot{TimesNet } &
    \rot{\textbf{\method}}
    \rule[0mm]{0mm}{20mm} 
    \\
    
    \hline
    \rowcolor{lightgray}
    Stream Processing  &\NO &\NO &\NO &\OK &\NO &\OK \rule[0mm]{0mm}{3.1mm} \\
    Forecasting        &\OK &\NO &\NO &\OK &\OK &\OK \rule[0mm]{0mm}{3.1mm} \\
    \rowcolor{lightgray} 
    Data Compression   &\NO &\OK &\NO &\OK &\NO &\OK \rule[0mm]{0mm}{3.1mm} \\
    Interdependency   &\NO &\OK &\OK &\NO &\NO &\OK \rule[0mm]{0mm}{3.1mm} \\
    \rowcolor{lightgray}
    \RELATION          &\NO &\NO &\NO &\NO &\NO &\OK \rule[0mm]{0mm}{3.1mm} \\
    \hline
    \end{tabular}
}
\normalsize
\vspace{1.0em}
\end{table}
 
}
In this section, we briefly describe investigations related to our work.
Table \ref{table:capability} summarizes the relative advantages of \method
with regard to
five
aspects.
\par
\myparaitemize{Time series modeling and forecasting}
Time series modeling and forecasting are important areas that
have attracted huge interest in many fields.
Autoregressive integrated moving average (ARIMA)~\cite{box1976arima} and Kalman filters (KF)~\cite{durbin2012time} are representative examples of traditional modeling and forecasting methods, and there have been many studies of their derivatives~\cite{
papadimitriou2003adaptive,
li2010parsimonious,
de2011forecasting,
shi2020block}.
TICC~\cite{hallac2017toeplitz} characterizes the interdependence
between different observations based on a Markov random field
but cannot capture the causal relationships.
Additionally, streaming algorithms have become more critical
in terms of processing a substantial amount of data under time/memory limitations,
and they have proved highly significant to the data mining and database community~\cite{
aggarwal2007data,
hahsler2016clustering,
matsubara2016regime,
kawabata2020non}. 
OrbitMap~\cite{matsubara2019dynamic} is the latest general method focusing on stream forecasting,
and it can find the transitions between major dynamic time series patterns.
However, it cannot discover the \relation between observations.
Research on deep learning models for forecasting has been very active in recent years~\cite{
Zeng2022AreTE,
Yuqietal-2023-PatchTST,
salinas2020deepar}.
TimesNet~\cite{wu2023timesnet} is a TCN-based method that transforms a 1D time series into 2D space based on multiple periods and captures complex temporal variations for forecasting.
Although deep learning-based methods are compelling,
their applicability to forecasting in a streaming fashion is limited
due to the prohibitively high computational costs associated with time series analysis,
which hinders continuous model updating with the most recent observations.
\par

\par
\myparaitemize{Causal inference/discovery}
Over decades,
a wide range of studies have been conducted on
causal inference/discovery~\cite{
shimizu2006linear,
he2021daring,
fujiwara2023causal,
jiang2023cfgode,
liu2023discovering} and 
addressing challenges based on the concept of causality~\cite{
richens2020improving,
dai2022graph,
wu2023causal}.
NOTEARS~\cite{zheng2018dags} is a new differentiable optimization framework for learning directed acyclic graphs, utilizing an acyclic regularization as a replacement for a combinatorial constraint.
Granger causality~\cite{granger1969investigating} has been widely used to analyze multivariate time series data.
However, Granger causality only indicates the presence of a predictive relationship~\cite{peters2017elements}.
Specifically, typical causality represents whether one observation causes another, while Granger causality represents whether one observation forecasts another~\cite{granger2014forecasting}.
In this paper, we focus on the cause-and-effect relationships that evolve over time in a data stream.
We try to discover them based on the structural equation model~\cite{pearl2009causality}, which is one of the most general formulations of causality.
\par
Consequently, none of these methods specifically 
focused on the discovery of the \relation
and forecasting future values in a streaming fashion, simultaneously.
\section{Proposed Model}
    \label{section:model}
    \TSK{
\begin{figure*}[t]
    \begin{tabular}{c}
        \begin{minipage}[b]{\linewidth}
            \centering
            \includegraphics[width=0.9\linewidth]{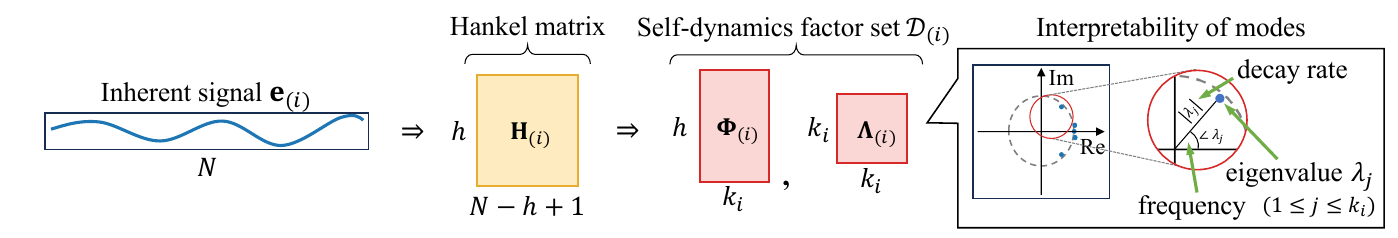}
            \vspace{-0.5em}
            \subcaption{Self-dynamics factor set (i.e., $\ith{\selfdynamics} = \{ \imode, \ieig \}$)}
            \label{fig:model:uni}
            \vspace{-0.4em}
        \end{minipage} \\
        \begin{minipage}[b]{\linewidth}
            \centering
            \includegraphics[width=0.9\linewidth]{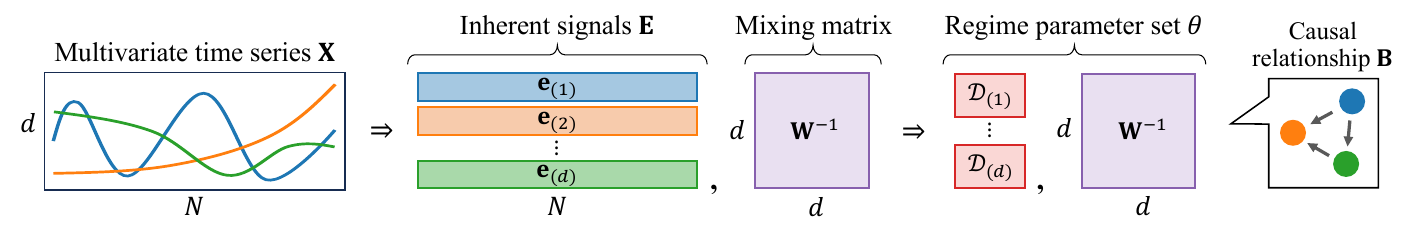}
            \vspace{-0.8em}
            \subcaption{Single regime parameter set (i.e., $\regime = \{ \demixing, \first{\selfdynamics}, ...,  \dth{\selfdynamics} \}$)}
            \label{fig:model:multi}
        \end{minipage}
    \end{tabular}
    \vspace{-1.4em}
    \caption{Illustration of \method:
    (a) we extract the latent temporal dynamics from the $i$-th univariate inherent signal $\ith{\ind}$, which behaves as a dynamical system.
    (b) The multivariate time series is described by mixing matrix $\demixing^{-1}$ and a collection of $d$ self-dynamics factor sets $\{ \first{\selfdynamics}, ..., \dth{\selfdynamics} \}$.
    The mixing matrix $\demixing^{-1}$ is not the same matrix as the causal adjacency matrix $\mB$,
    it is instrumental in identifying the \relation.}
    \label{fig:model}
    \vspace{-1.0em}
\end{figure*}
}
\TSK{
\begin{table}[t]
\vspace{-0.5em}
\centering
\small
\caption{Symbols and definitions.}
\label{table:define}
\vspace{-1.2em}
\begin{tabular}{l|l}
\toprule
Symbol & Definition \\
\midrule
$d$ & Number of dimensions \\
$t_c$ & Current time point \\
$\mX$ & Co-evolving multivariate data stream (semi-infinite) \\
$\mX^c$ & Current window, i.e., $\mX^c = \mX[t_m:t_c]\in\R^{d\times N}$ \\
\midrule
$h$ & Embedding dimension \\
$\embed{\cdot}$ & Observable for time-delay embedding, i.e., $g\colon\R \rightarrow \R^{h}$ \\
$\mH$ & Hankel matrix \\
$\nmodes$ & Number of modes \\
$\modes$ & Modes of the system, i.e., $\modes \in \R^{h\times\nmodes}$ \\
$\eigs$ & Eigenvalues of the system, i.e., $\eigs \in \R^{\nmodes\times\nmodes}$ \\
$\demixing$ & Demixing matrix, i.e., $\mW = [\rowvect{w}_1, ..., \rowvect{w}_d]^\top \in \R^{d \times d}$ \\
$\mB$ & Causal adjacency matrix, i.e., $\mB \in \R^{d \times d}$ \\
\midrule
$\ind(t)$ & Inherent signal at time point $t$, i.e., $\ind(t) = \{ \ith{e}(t) \}_{i=1}^d$ \\
$\mat{S}(t)$ & Latent vectors at time point $t$, i.e., $\mat{S}(t) = \{ 
\ith{\vs}(t) \}_{i=1}^d$ \\
$\vvec(t)$ & Estimated vector at time point $t$, i.e., $\vvec(t) = \{ \ith{v}(t) \}_{i=1}^d$ \\
\midrule
$\mathcal{D}$ & Self-dynamics factor set, i.e., $\mathcal{D} = \{\modes, \eigs\}$\\
$\regime$ & Regime parameter set, i.e., $\regime = \{ \mW, \mathcal{D}_{(1)}, ..., \mathcal{D}_{(d)} \}$\\
\midrule
$R$ & Number of regimes \\
$\regimeset$ & Regime set, i.e., $\regimeset 
 = \{ \regime^1, ..., \regime^R \}$\\
 $\mathcal{B}$ & \Relation, i.e., $\mathcal{B} = \{\mB^1, ..., \mB^R\}$\\
$\updateset$ & Update parameter set,  i.e., $\updateset 
 = \{ \update^1, ..., \update^R \}$ \\
\midrule
 $\modelparam$ & Full parameter set,  i.e., $\modelparam 
 = \{ \regimeset, \updateset \}$ \\

\bottomrule
\end{tabular}
\normalsize
\vspace{1.0em}
\end{table}

}
In this section, we present our proposed model.
The main symbols we use in this paper are described in
Table \ref{table:define}.
Here,
before introducing the main topic,
we briefly describe the principles and concepts of \method.
We design our proposed model based on the structural equation model (SEM)~\cite{pearl2009causality},
which is written as $\mX_{\text{sem}} = \mB_{\text{sem}}\mX_{\text{sem}} + \mE_{\text{sem}}$,
where $\mX_{\text{sem}}$ is the observed variables,
$\mB_{\text{sem}}$ is the causal adjacency matrix, and
$\mE_{\text{sem}}$ is a set of mutually independent exogenous variables with a non-Gaussian distribution.
Note that we assume that
the data generating process is linear,
the causal network is a directed acyclic graph, and
there are no unobserved confounders in this paper.
The structural equation model can express a typical causality,
however, in real-world applications,
causal relationships change over time
in accordance with the transitions of distinct dynamical patterns.
In our model, we assume that the exogenous variables behave as a dynamical system;
however, it is inappropriate to consider
their time-evolution
as a single dynamical system
due to their independence from each other.
In summary, given a multivariate data stream
which contains various distinct dynamical patterns (i.e., regimes),
we aim to discover the \relation and summarize an entire data stream on the above assumption.
Specifically, we need to capture the following properties
to achieve
the above objective:
\begin{itemize}
    \item [\textbf{(P1)}] latent temporal dynamics of exogenous variables
    \item [\textbf{(P2)}] dynamical pattern in a single regime
\end{itemize}
So, how can we build our model that expresses both \textbf{(P1)} and \textbf{(P2)}?
What is the acceptable mathematical model that summarizes a data stream and discovers the \relation?
To handle \textbf{(P1)}, we express each of the exogenous variables as the superposition of computed basis vectors (i.e., modes).
We model \textbf{(P2)} by combining the above components.
We provide our answers below.
\subsection{Proposed Solution: \method}
We now present our model in detail.
First, we provide the definition for our proposed method.
\begin{definition}[Inherent signals: $\mE$]
    Let $\mE$ be a bundle of $d$ mutually independent signals with a non-Gaussian distribution, i.e., $\mE = \{ \ith{\ind} \}_{i=1}^d$, where $\ith{\ind} = \{ \ith{e}(1), ..., \ith{e}(t) \}$ is the $i$-th univariate inherent signal. The main property is that they evolve over time.
\end{definition}
\noindent Figure \ref{fig:model} is an illustration of our proposed model.
In the first half of this section,
we describe \textbf{(P1)} the latent temporal dynamics of the $i$-th univariate inherent signal $\ith{\ind}$
by introducing the self-dynamics factor set $\ith{\selfdynamics}$,
and next, we propose the parameter set $\regime$ to represent \textbf{(P2)} regimes and an entire data stream.
\subsubsection{Latent temporal dynamics of an inherent signal (P1)}
\label{section:uni}
First, we answer the fundamental question, namely,
how can we extract the latent temporal dynamics from the $i$-th inherent signal (i.e., $\ith{\ind} = \{ \ith{e}(1), ..., \ith{e}(t) \}$) and express it as a superposition of the modes?
The difficulty arises from the fact that
the latent dynamics in the system are generally multi-dimensional,
making a single dimension inadequate for modeling the system.
Here, we utilize state space augmentation methods
to compensate for this inadequacy.
In particular, we adopt time-delay embedding,
which is effective in capturing nonlinear dynamics.
Specifically, this is an established method
for the geometric reconstruction of attractors
for nonlinear systems
based on the measurement of generic observables,
$\embed{\ith{e}(t)} \coloneqq (\ith{e}(t), \ith{e}(t-1), ..., \ith{e}(t-h+1)) \in \R^{h}$,
where $h$ is the embedding dimension.
We form the Hankel matrix $\ith{\hankel} \in \R^{h \times (t-h+1)}$
by using the above observable $\embed{\cdot}$.
\begin{align}
    \label{eq:timedelay}
    \ith{\hankel} &= 
    \begin{bmatrix}
        | & | &  & | \\
        \embed{\ith{e}(h)} & \embed{\ith{e}(h+1)} & \cdots & \embed{\ith{e}(t)} \\
        | & | &  & |
    \end{bmatrix}
\end{align}
As seen in Eq. \eqref{eq:timedelay},
each state represented by a single measurement function is augmented with its past history.
Furthermore, according to Takens' embedding theorem~\cite{takens2006detecting},
it is guaranteed that time-delay embedding produces a vector
whose dynamics are diffeomorphic to the dynamics of the original state under certain conditions.
Intuitively,
the reconstruction theoretically preserves the properties of the original dynamical system,
allowing an analysis of Hankel matrix to reveal important features that may not be directly apparent in the original data alone.
In many cases, an embedding dimension may be chosen without sacrificing the diffeomorphism. \par
We now extract the latent temporal dynamics
expressed as the superposition of the modes
from the $i$-th univariate inherent signal $\ith{\ind}$
using the above method.
We thus introduce a time-evolving activity for describing the dynamical system of the $i$-th inherent signal $\ith{\ind}$.
This activity is a latent vector $\ith{\latent}(t)\in\C^{\nmodes_i}$, which is $\nmodes_i$-dimensional complex-valued latent vector at time point $t$, where $\nmodes_i$ is the number of modes.
Consequently,
the dynamical system for the $i$-th univariate inherent signal $\ith{\ind}$
can be described with the following equations:
\begin{model}
\label{model:uni}
Let $\ith{\latent}(t)$ be the $\nmodes_i$-dimensional latent vector at time point $t$ for $i\in\{1, \ldots, d\}$.
The following equations govern the $i$-th univariate inherent signal $\ith{\ind}$,
\begin{align}
    \label{eq:univariate}
    \begin{split}
        \ith{\latent}(t + 1) &= \ieig\ith{\latent}(t) \\
        \ith{e}(t) &= \invembed{\imode\ith{\latent}(t)}
    \end{split}
    \normalsize
\end{align}
where $\invembed{\cdot}:\R^h\to\R$ is the inverse of the observables $\embed{\cdot}$,
each column of $\imode$ is one mode, and $\ieig$ is a diagonal matrix containing the $\nmodes_i$ eigenvalues corresponding to those modes.
\end{model}
\noindent Note that the latent vector $\ith{\latent}(t)$ is expressed as a superposition of $\nmodes_i$ modes.
The eigenvalues $\ieig\in\C^{\nmodes_i\times\nmodes_i}$ describe latent dynamical activities, and
the modes $\imode\in\C^{h\times\nmodes_i}$ and $\invembed{\cdot}$ are the observation projections
that generate the $i$-th univariate inherent signal $\ith{e}(t)$ at each time point $t$.
Consequently, we have the following:

\begin{definition}[Self-dynamics factor set: $\ith{\selfdynamics}$]
    Let $\ith{\selfdynamics}$ be a set of modes $\imode$ and eigenvalues $\ieig$,
    i.e., $\ith{\selfdynamics} = \{ \imode, \ieig \}$,
    which represents the latent temporal dynamics of the $i$-th univariate inherent signal $\ith{e}$.
\end{definition}

\noindent We can interpret the features of the above model by focusing on the self-dynamics factor set $\ith{\selfdynamics}$.
Specifically,
the eigenvalues $\ieig$ imply the temporal dynamics of the modes $\imode$,
such as exponential growth/decay and oscillations.
These are derived from a characteristic of a discrete dynamical system.
Considering the eigenvalues $\ieig$ represent
the behavior of a discrete dynamical system with sampling interval $\Delta t$, decay rate $a$ and temporal frequency $b$ of the $j$-th mode $\varphi_j$ are shown as follows, using the $j$-th eigenvalue $\lambda_j$:
\begin{align*}
a = \frac{\Re (\log{\lambda_j})}{\Delta t}, \quad b = \frac{\Im (\log{\lambda_j})}{\Delta t}
\end{align*}
where $\Re$ and $\Im$ are the real and imaginary parts, respectively.
In addition, note that $\log{\lambda_j} = \ln{|\lambda_j|} + i\arg{\lambda_j}$, and
it can be said that the decay rates and temporal frequencies of the modes are given by the absolute values and arguments of the eigenvalues, respectively.
\subsubsection{Dynamical pattern in a single regime (P2)}
Thus, we have seen how to model the latent temporal dynamics of the $i$-th univariate inherent signal $\ith{\ind}$
using self-dynamics factor set $\ith{\selfdynamics}$.
Here, let us tackle the next question, namely
how can we describe
the major dynamical pattern (i.e., regime)
considering the \relation between observations in a data stream?
\noindent
We establish a model to combine the $d$ self-dynamics factor sets (i.e., $\first{\selfdynamics}, ..., \dth{\selfdynamics}$) for generating the estimated vector $\est(t)\in\R^d$ at time point $t$.
Also, we need a set of $d$ latent vectors (i.e., $\mat{S}(t) = \{ \ith{\latent}(t) \}_{i=1}^d$).
Consequently, we extend Model \ref{model:uni}, and the dynamical system for $d$-dimensional multivariate time series can be described with the following equations:

\begin{model}
    \label{model:multi}
    Let $\ith{\latent}(t)$ be the $\nmodes_i$-dimensional latent vector for the $i$-th univariate inherent signal $\ith{e}(t)$ at time point $t$,
    $\ind(t)$ be the $d$-dimensional inherent signals at time point $t$ (i.e., $\ind(t) = \{ \ith{e}(t) \}_{i=1}^d$), and
    $\est(t)$ be the $d$-dimensional estimated vector at time point $t$.
    The following equations govern the single regime,
    \begin{align}
        \label{eq:multivariate}
        \begin{split}
            \ith{\latent}(t + 1) &= \ieig\ith{\latent}(t) \quad (1 \leq i \leq d) \\
            \ith{e}(t) &= \invembed{\imode\ith{\latent}(t)} \quad (1 \leq i \leq d) \\
            \vect{v}(t) &= \demixing^{-1}\ind(t)
        \end{split}
    \end{align}
\end{model}

\noindent In Model \ref{model:multi},
we require an additional parameter, demixing matrix $\demixing$,
which represents the relationships among $d$ inherent signals (i.e., $\first{\ind}, ..., \dth{\ind}$) and
is instrumental in identifying the \relation in data streams.
However, there are indeterminacies in a mixing matrix  (i.e., the inverse of a demixing matrix),
so it is not the same matrix as the causal adjacency matrix $\mB$. We present the algorithm for obtaining $\mB$ from $\demixing$ in Section \ref{section:alg:generator}.
Consequently, we have the following:
\begin{definition}[Single regime parameter set: $\regime$]
Let $\regime$ be a parameter set of a single regime, i.e., $\regime = \{ \demixing, \first{\selfdynamics}, ..., \dth{\selfdynamics} \}$, where $\demixing$ serves as the basis for generating the causal adjacency matrix $\mB$.
\end{definition}
\noindent Furthermore, we want to detect the transitions of regimes, which induce changes in causal relationships.
Let $R$ denote the optimal number of regimes up to the time point $t$.
Then, a data stream $\mX$ is summarized using a set of $R$ regimes (i.e., $\regime^1, ..., \regime^R$).
Consequently, a regime set for a data stream $\mX$ and \relation are defined as follows:
\begin{definition}[Regime set: $\regimeset$]
Let $\regimeset$ be a parameter set of multiple regimes, i.e.,$\regimeset = \{ \regime^1, ..., \regime^R \}$, which describes multiple distinct dynamical patterns in an entire data stream.
\end{definition}
\begin{definition}[\Relation: $\mathcal{B}$]
    Let $\mathcal{B}$ be a parameter set of causal adjacency matrices, i.e., $\mathcal{B} = \{ \mB^1, ..., \mB^R \}$, which describes time-changing of causal relationships. Note that $\mB^i$ is a causal adjacency matrix corresponding to the $i$-th regime $\regime^i$.
\end{definition}

\section{Optimization Algorithm}
    \label{section:algorithm}
    \TSK{
\begin{figure}[t]
    \begin{center}
        \includegraphics[width=1.05\linewidth]{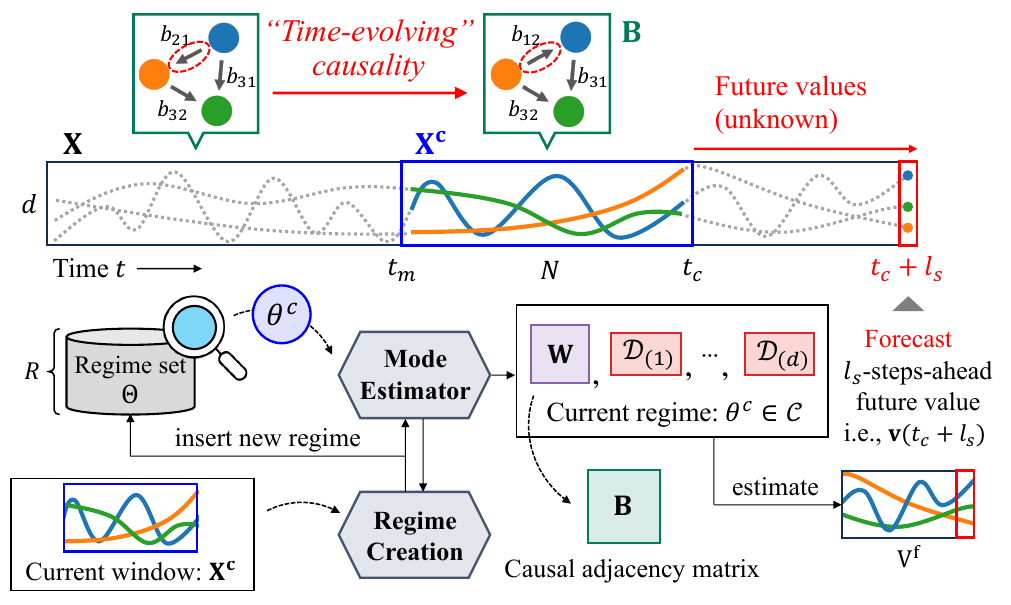}
        \vspace{-4.0ex}
        \caption{Overview of \method algorithm: Given a data stream $\mX$, it performs all the following tasks at every time point $t_c$. Firstly, it searches for the best regime $\regime^c$ for the current window $\mX^c$. It then forecasts the $l_s$-steps-ahead future value, i.e., $\est(t_c+l_s)$ by utilizing the best one. When the algorithm encounters an unknown pattern in $\mX^c$, it estimates a new regime $\regime$ and inserts it into $\regimeset$.}
        \label{fig:algorithm}
    \end{center}
    \vspace{-0.5ex}
\end{figure}
}
Thus far, we have shown how we describe the major dynamical pattern using the demixing matrix $\demixing$ which can be transformed into a causal adjacency matrix $\mB$.
In this section, we present an effective algorithm
to identify the \relation $\mathcal{B}$ and
estimate the regime set $\regimeset$.
Figure \ref{fig:algorithm} shows an overview of our proposed algorithm.
We first present an effective way to estimate a new model parameter set from a multivariate time series, where we assume it has only a single regime.
We then describe a streaming algorithm to identify $\mathcal{B}$ and maintain $\regimeset$ incrementally for multiple distinct 
dynamical patterns, simultaneously.
\subsection{\textsc{RegimeCreation}}
\label{section:alg:creation}
We first propose an algorithm, namely \textsc{RegimeCreation}, for estimating a single regime parameter set $\theta = \{ \demixing, \first{\selfdynamics}, ..., \dth{\selfdynamics} \}$ from a data stream $\mX$.
The algorithm consists of two main steps: (i) it decomposes $\mX$ into a demixing matrix $\demixing$ and inherent signals $\mE$ and (ii) it computes $d$ self-dynamics factor sets $\{ \first{\selfdynamics}, ..., \dth{\selfdynamics} \}$ according to Eq. \eqref{eq:univariate}.
We use ICA for the decomposition of $\mX$, where it is essential for identifying the optimal causality.
Next, as regards computing the $i$-th self-dynamics factor set $\ith{\selfdynamics}$, assume that we have the following data matrices based on the Hankel matrix $\ith{\hankel}$:
\begin{align*}
    \ith{\mat{L}} = 
    \begin{bmatrix}
    \embed{\ith{e}(h+1)} & \cdots & \embed{\ith{e}(t)}
    \end{bmatrix} \in \R^{h \times (t-h)} \\
    \ith{\mat{R}} = 
    \begin{bmatrix}
    \embed{\ith{e}(h)} & \cdots & \embed{\ith{e}(t-1)}
    \end{bmatrix} \in \R^{h \times (t-h)}
\end{align*}
And, we use the following weight cost function:
\begin{align}
    \begin{split}
    \label{eq:loss}
    & \min_{\trans}\sum_{t'=h}^{t-1}\forgetting^{2(t-1-t')}||\embed{\ith{e}(t'+1)} - \ith{\trans}\embed{\ith{e}(t')}||_2^2 \\
    = & \min_{\trans} || (\ith{\mat{L}}-\ith{\trans}\ith{\mat{R}})\Forgetting ||_F^2
    \end{split}
\end{align}
where, $\ith{\trans}$ is the transition matrix, from which the eigendecomposition yields the modes $\imode$ and the corresponding eigenvalues $\ieig$, and $\Forgetting = \mathrm{diag}(\forgetting^{t-h-1}, ..., \forgetting^{0}) \in \mathbb{R}^{(t-h)\times(t-h)}$ is defined as a forgetting matrix, based on the recursive least squares principle.
In addition, according to Koopman theory~\cite{koopman1931hamiltonian}, while the transition matrix $\ith{\trans}$ is a linear operator,
it is applicable even to nonlinear
dynamical systems, unlike the classical modal decomposition of linear time-invariant systems.
Specifically, the analytical algorithm proceeds as follows:

{\setlength{\leftmargini}{15pt}
\begin{enumerate}
    \renewcommand{\labelenumi}{\Roman{enumi}.}
    \item Compute the ICA of $\mX = \demixing^{-1}\mE$.
    \item Form the Hankel matrix $\ith{\hankel}$
    according to Eq. \eqref{eq:timedelay}.
    \item Build a pair of data matrices $(\ith{\mat{L}}, \ith{\mat{R}})$.
    \item Compute the SVD of $\ith{\mat{R}}\Forgetting = \ith{\mat{U}}\ith{\boldsymbol{\Sigma}}{\ith[\top]{\mat{V}}}$.
    We automatically determine the optimal number of singular values $\nmodes_i$ by~\cite{gavish2014optimal}.
    \item Project the transition matrix $\ith{\trans}$
    onto the $\nmodes_i$-dimensional subspace spanned by the left singular vector $\mat{U}^{(i)}$.
    \begin{align*}
        \ith{\tilde{\trans}} = \ith[\top]{\mat{U}}\ith{\trans}\ith{\mat{U}} = {\ith[\top]{\mat{U}}}\ith{\mat{L}}\Forgetting\ith{\mat{V}}\ith{\mat{\Sigma}}^{-1} \in \R^{\nmodes_i \times \nmodes_i}
    \end{align*}
    \item Compute the eigendecomposition of $\ith{\tilde{\trans}}\ith{\mat{Z}} = \ith{\mat{Z}}\ieig$.
    Note that the eigenvalues $\ieig$
    are identical to the $\nmodes_i$ leading eigenvalues of $\ith{\trans}$
    because the left singular vector $\ith{\mat{U}}$ is an orthogonal matrix.
    \item Compute the modes $\imode = \ith{\mat{U}}\ith{\mat{Z}}$.
\end{enumerate}
}
\begin{lemma}[Time complexity of \textsc{RegimeCreation}]
\label{lemma:create_time}
The time complexity of \textsc{RegimeCreation} is $O(N(d^2+h^2)+k^3)$, where $k=\max_i(k_i)$.
Please see Appendix\,\ref{section:app:algorithm} for details.
\end{lemma}
\subsection{Streaming Algorithm}
Our next step is to answer the most important question:
how can we employ our proposed model
for identifying the causal adjacency matrix $\mB$ from the demixing matrix $\demixing\in\regime$ and 
forecasting future values in a streaming fashion?
Before turning to the main topic,
we provide the definitions of some key concepts.
\begin{definition}[Update parameter: $\update$] Let $\update$ be a parameter set for updating a regime $\theta$,
i.e., $\update = \{ \{ \ith{\bm{P}} \}_{i=1}^d, \{ \ith{\bm{\epsilon}} \}_{i=1}^d \}$,
where
$\ith{\bm{P}} = 
(\ith{\mat{R}}\Forgetting{\ith[\top]{\mat{R}}})^{-1}$ and $\ith{\bm{\epsilon}}$ is the energy.
\end{definition}
\begin{definition}[Full parameter set: $\modelparam$] Let $\modelparam$ be a full parameter set of \method,
i.e., $\modelparam = \{ \regimeset, \updateset \}$,
where $\regimeset$ and $\updateset$ consist of $R$ regimes and update parameters, respectively,
namely, $\regimeset = \{ \regime^1, ..., \regime^R \}$,
and $\updateset = \{ \update^1, ..., \update^R \}$.
\end{definition}
\noindent With the above definitions, the formal problem is as follows:
\begin{problem}
\textbf{Given} a multivariate data stream $\mX$,
where $\vx(t_c)$ is the most recent value at time point $t_c$,
\begin{itemize}
    \item \textbf{Find} the optimal full parameter set, i.e., $\modelparam = \{ \regimeset, \updateset \}$,
    \item \textbf{Discover} the \relation, i.e., $\mathcal{B}$,
    \item \textbf{Forecast} an $l_s$-steps-ahead future value, i.e., $\vvec(t_c+l_s)$.
\end{itemize}
\end{problem}
\noindent Here, we refer to the regime for the current window $\mX^c = \mX[t_m:t_c]$ as $\regime^c$, and the update parameter corresponding to $\regime^c$ as $\update^c$. In addition, we need the latent vectors $\mat{S}(t_c)$ at the current time $t_c$
for forecasting an $l_s$-steps-ahead future value $\vvec(t_c+l_s)$, and so keep it as $\mat{S}^c_{en}$.
In summary, our proposed algorithm keeps them as the model candidate $\candparam = \{ 
\regime^c, \update^c, \mat{S}^c_{en} \}$ for stream processing.
\subsubsection{Overview}
We now introduce our streaming algorithm, \method, which consists of the following algorithms.
For detailed descriptions and pseudocode, we would refer you to Algorithm \ref{alg:model} in Appendix \ref{section:app:algorithm}.
{\setlength{\leftmargini}{15pt}
\begin{itemize}
    \item \modelestimator:
    Estimates the optimal full parameter set $\modelparam$ and the model candidate $\candparam$ which appropriately describes the current window $\mX^c$.
    \item \modelgenerator:
    Forecasts an $l_s$-steps-ahead future value, i.e., $\vvec(t_c+l_s)$, and identifies the causal adjacency matrix $\mat{B}$, using the model candidate $\candparam$.
    \item \regimeupdate:
    Updates the current regime $\regime^c$ using update parameter $\update^c\in\candparam$ and the most recent value $\vx(t_c)$. This is only performed if a new regime is not created.
\end{itemize}
}
\noindent
The following sections provide detailed explanations.
\subsubsection{\modelestimator}
Given a new value $\vx(t_c)$ at the current time $t_c$, we first need to update incrementally the full parameter set $\modelparam$ and the model candidate $\candparam$, which best describes the current window $\mX^c$.
Algorithm \ref{alg:estimator} (See Appendix \ref{section:app:algorithm}) is the \modelestimator algorithm in detail.
Here, let $f(\mX^c; \mat{S}_0^c, \regime^c)$ be a new function
for estimating the optimal parameter
so that it minimizes the mean square errors
between the current window $\mX^c$ and the estimated window $\mat{V}^c$ in Model \ref{model:multi},
i.e., $f(\mX^c; \mat{S}_0^c, \regime^c) = \sum_{t=t_m+h-1}^{t_c}||\bm{x}(t) - \bm{v}(t)||$,
where $\mat{S}_0^c$ represents the latent vectors at time point $t_m+h-1$.
Note that when embedding the time series using $\embed{\cdot}$,
the number of data points (namely, the number of columns in the Hankel matrix $\hankel$) 
is partially reduced compared with before embedding.
The most straightforward way to determine $\mat{S}_0^c$
is to adopt $\{ {\modes_{(i)}^\dagger}\embed{\ith{e}(t_m + h - 1)}\}_{i=1}^d$
according to Eq. \eqref{eq:multivariate}.
However, the noisy initial conditions give rise to unexpected forecasting.
Therefore, we optimize $\mat{S}_0^c$ by using the Levenberg-Marquardt (LM) algorithm~\cite{more2006levenberg}
and thus enable the effects of noise in observations to be removed.
Here, we return to the \modelestimator algorithm, which proceeds as follows:
{\setlength{\leftmargini}{15pt}
\begin{enumerate}
    \renewcommand{\labelenumi}{\Roman{enumi}.}
    \item It optimizes initial condition $\mat{S}_0^c$,
    so that it minimizes the errors between the current window $\mX^c$ and the current regime $\regime^c$.
    \item If $f(\mX^c; \mat{S}_0^c, \regime^c) > \tau$, it searches for a better regime $\regime \in \regimeset$.
    \item If $f(\mX^c; \mat{S}_0^c, \regime^c) > \tau$ still holds, it creates a new regime for $\mX^c$ using \textsc{RegimeCreation}, and inserts it into $\regimeset$.
\end{enumerate}
}
\par
\par
\subsubsection{\modelgenerator}
\label{section:alg:generator}
The next algorithm is \modelgenerator,
which incrementally identifies the causal adjacency matrix $\mB$ and
forecasts an $l_s$-steps-ahead future value $\est(t_c+l_s)$
by using the model candidate $\candparam$.
As for forecasting,
it generates the value of $\est(t_c+l_s)$ according to Eq. \eqref{eq:multivariate}
with the most suitable regime $\regime^c$ for $\mX^c$,
which is selected by $\modelestimator$.
On the other hand,
we identify the causal adjacency matrix $\mB$ from the demixing matrix $\demixing \in \regime^c$.
A mixing matrix (i.e., the inverse of a demixing matrix) typically has the two major indeterminacies:
the order and scaling of the independent components;
however, we must address the above difficulties if we are to identify the optimal causal adjacency matrix~\cite{shimizu2006linear}.
Consequently, the algorithm for identifying the causal adjacency matrix $\mB$ proceeds as follows:
{\setlength{\leftmargini}{15pt}
\begin{enumerate}
\renewcommand{\labelenumi}{\Roman{enumi}.}
\item Find the permutation of rows of $\demixing$ that yields a matrix $\tilde{\demixing}$ without any zeros on the main diagonal.
\item Divide each row of $\tilde{\demixing}$ by its corresponding diagonal element to yield a new matrix $\tilde{\demixing'}$ with all ones on the diagonal.
\item Compute an estimate $\hat{\mB}$ of $\mB$ using $\hat{\mB} = \mat{I} - \tilde{\demixing'}$.
\item Finally, to find a causal order, compute the permutation matrix $\mat{K}$ of $\hat{\mB}$ that yields a matrix $\tilde{\mat{B}} = \mat{K}\hat{\mB}\mat{K}^\top$, which minimizes the sum of the elements in the upper triangular part of $\tilde{\mat{B}}$.
\end{enumerate}
}
\noindent This algorithm resolves two major indeterminacies in a mixing matrix in steps I and II.
Moreover, it finds the causal order, in other words, it removes the insufficient connection in step IV.
The causal relationships are already identified up to step III, but this step is important for visualizing the resulting directed acyclic graph.
\begin{lemma}[Causal identifiability]
\label{lemma:causal}
Causal discovery in \method is equivalent to
finding the causal adjacency matrix $\mB$ in \modelgenerator.
Please see Appendix\,\ref{section:app:algorithm} for details.
\end{lemma}
\noindent This lemma demonstrates theoretically that our proposed algorithm is capable of discovering causal relationships.
\par
\subsubsection{\regimeupdate}
\label{section:alg:update}
Finally, when an existing regime is selected as the current regime $\regime^c$ from the regime set $\regimeset$,
we update its parameters (i.e., $\demixing, \selfdynamics^{(1)}, ..., \selfdynamics^{(d)}$) using a new value $\vx(t_c)$
to ensure that this regime represents a more sophisticated dynamical pattern. 
In short, \regimeupdate has two parts:
(i) update the demixing matrix $\demixing$ and 
(ii) update each self-dynamics factor set $\ith{\selfdynamics}$.
In part (i),
we use an algorithm based on adaptive filtering techniques~\cite{yang1995projection, haykin2002adaptive}.
It is so efficient in terms of computational and memory requirements,
while converging quickly,
with no special parameters to tune.
Here, we briefly describe the update process for $\demixing$:
{\setlength{\leftmargini}{15pt}
\begin{enumerate}
    \renewcommand{\labelenumi}{\Roman{enumi}.}
    \item Compute the $i$-th univariate inherent signal $\embed{\ith{e}(t_c)}$ at the current time $t_c$, by projecting $\vx(t_c)$ onto $\rowvect{\vect{w}}_i$, which is the $i$-th row vector of $\demixing$, before the update.
    \item Estimate the reconstruction error and the energy $\ith{\bm{\epsilon}}$ based on the value of $\embed{\ith{e}(t_c)}$, 
    \item Update the estimates of $\rowvect{\vect{w}}_i$ using error and energy $\ith{\bm{\epsilon}}$.
\end{enumerate}}
\noindent In part (ii),
we update the self-dynamics factor set $\ith{\selfdynamics}$ by utilizing the following recurrence:
\begin{align}
    \label{eq:update_trans}
    \begin{split}
        \ith[new]{\trans} &= \ith[prev]{\trans} + (\embed{\ith{e}(t_c)}-\ith[prev]{\trans}\embed{\ith{e}(t_c-1)})\ith{\boldsymbol\gamma} \\
        \ith{\boldsymbol\gamma} &= \frac{\embed{\ith{e}(t_c-1)}^\top\ith[prev]{\bm{P}}}{\forgetting + \embed{\ith{e}(t_c-1)}^\top\ith[prev]{\bm{P}}\,\embed{\ith{e}(t_c-1)}} \\
        \ith[new]{\bm{P}} &= \frac{1}{\forgetting}(\ith[prev]{\bm{P}} - \ith[prev]{\bm{P}}\,\embed{\ith{e}(t_c-1)}\ith{\boldsymbol\gamma})
    \end{split}
\end{align}
In this equation, $\imode$ and $\ieig$ are eigenvectors and eigenvalues of $\ith{\trans}$, respectively.
This recurrence minimizes the weighted cost function as outlined in Eq. \eqref{eq:loss}, focusing on a new value $\vx(t_c)$, thereby adapting to most recent patterns.
See Appendix \ref{section:app:algorithm} for details regarding the above formula.
In summary, \regimeupdate proceeds in the following manner:
{\setlength{\leftmargini}{15pt}
\begin{enumerate}
    \renewcommand{\labelenumi}{\Roman{enumi}.}
    \item In accordance with the part (i) algorithm, update the demixing matrix $\demixing$ using a new value $\vx(t_c)$.
    \item Compute the current inherent signals $\mE^c$ from the current window $\mX^c$ using the updated demixing matrix $\mW$.
    \item Update each self-dynamics factor set $\ith{\selfdynamics}$ according to Eq. \eqref{eq:update_trans}.
\end{enumerate}}
\par
\begin{lemma}[Time complexity of \method]
\label{lemma:stream_time}
Based on Lemma \ref{lemma:create_time}, the time complexity of \method is
at least $O(N\sum_ik_i+dh^2)$,
and at most $O(RN\sum_i k_i+N(d^2+h^2)+k^3)$ per process.
Please see Appendix\,\ref{section:app:algorithm} for details.
\end{lemma}
\noindent This theoretical analysis indicates that
our proposed algorithm requires only constant computational time
with regard to the entire data stream length $t_c$.
Therefore, \method is practical for semi-infinite data streams in terms of execution speed.

\section{Experiments}
    \label{section:experiments}
\TSK{
\begin{table*}[t]
    \centering
    \caption{
    Causal discovering results with multiple temporal sequences to encompass various types of real-world scenarios.
    }
    \vspace{-1.0em}
    \begin{tabular}{>{\centering}p{4.66em}|cc|cc|cc|cc|cc|cc|cc|cc}
        \toprule
        Models
        & \multicolumn{2}{c|}{\method }
        & \multicolumn{2}{c|}{CASPER }
        & \multicolumn{2}{c|}{DARING }
        & \multicolumn{2}{c|}{NoCurl }
        & \multicolumn{2}{c|}{NO-MLP }
        & \multicolumn{2}{c|}{NOTEARS }
        & \multicolumn{2}{c|}{LiNGAM }
        & \multicolumn{2}{c}{GES}
        \\
        \midrule 
        Metrics
        & \,SHD & SID\,
        & \,SHD & SID\,
        & \,SHD & SID\,
        & \,SHD & SID\,
        & \,SHD & SID\,
        & \,SHD & SID\,
        & \,SHD & SID\,
        & \,SHD & SID\,
        \\
        \midrule
        $1,2,1$ & \,\textbf{3.82} & \textbf{4.94}\, & \,5.58 & \underline{7.25}\, & \,5.75 & 8.58\, & \,6.31 & 9.90\, & \,6.36 & 8.74\, & \,\underline{5.03} & 9.95\, & \,7.13 & 8.23\, & \,7.49 & 11.7\, \\
        $1,2,3$ & \,\textbf{4.48} & \textbf{6.51}\, & \,5.97 & 8.44\, & \,5.81 & 9.17\, & \,6.13 & 9.51\, & \,6.44 & 8.77\, & \,\underline{5.69} & 9.56\, & \,6.79 & \underline{7.33}\, & \,7.03 & 10.1\, \\
        $1,2,2,1$ & \,\textbf{4.32} & \textbf{5.88}\, & 5.41 & \underline{8.41}\, & \,6.54 & 9.17\, & \,6.69 & 10.0\, & \,6.55 & 8.72\, & \,\underline{5.23} & 9.54\, & \,7.12 & 8.65\, & \,7.08 & 9.77\, \\
        $1,2,3,4$ & \,\textbf{4.21} & \textbf{5.76}\, & \,6.22 & \underline{8.33}\, & \,6.12 & 9.58\, & \,6.10 & 9.61\, & \,6.62 & 8.87\, & \,\underline{5.73} & 10.1\, & \,7.10 & 8.50\, & \,7.29 & 11.3\, \\
        $1,2,3,2,1$ & \,\textbf{4.50} & \textbf{6.11}\, & \,6.02 & 8.28\, & \,\underline{5.45} & \underline{7.77}\, & \,6.20 & 9.83\, & \,6.56 & 8.83\, & \,{5.57} & 9.11\, & \,7.46 & 8.05\, & \,7.74 & 12.1\, \\
        \bottomrule
    \end{tabular}
    \label{table:discovering_accuracy}
    \vspace{-0.5em}
\end{table*}
 
}
\TSK{
\begin{table*}[t]
    \centering
    \caption{
    Multivariate forecasting results for both synthetic and real-world datasets. We used forecasting steps $l_s\in\{5, 10, 15\}$.
    }
    \vspace{-1.0em}
    \begin{tabular}{c|c|cc|cc|cc|cc|cc|cc}
        \toprule
        \multicolumn{2}{c|}{Models}
        & \multicolumn{2}{c|}{\method }
        & \multicolumn{2}{c|}{TimesNet }
        & \multicolumn{2}{c|}{PatchTST }
        & \multicolumn{2}{c|}{DeepAR }
        & \multicolumn{2}{c|}{OrbitMap }
        & \multicolumn{2}{c}{ARIMA }
        \\
        \midrule 
        \multicolumn{2}{c|}{Metrics}
        & RMSE & MAE
        & RMSE & MAE
        & RMSE & MAE
        & RMSE & MAE
        & RMSE & MAE
        & RMSE & MAE
        \\
        \midrule
        \multirow[t]{3}{*}{\synthetic}
        & 5 
        & \textbf{0.722} & \textbf{0.528}
        & 0.805 & {0.578} 
        & \underline{0.768} & 0.581 
        & 1.043 & 0.821 
        & 0.826 & \underline{0.567} 
        & 0.962 & 0.748 \\       
        & 10 
        & \textbf{0.829} & \textbf{0.607}
        & \underline{0.862} & 0.655 
        & 0.898 & {0.649} 
        & 1.073 & 0.849 
        & 0.896 & \underline{0.646} 
        & 0.966 & 0.752 \\
        & 15 
        & \textbf{0.923} & \textbf{0.686}
        & \underline{0.940} & \underline{0.699} 
        & 0.973 & 0.706 
        & 1.137 & 0.854 
        & 0.966 & 0.710 
        & 0.982 & 0.765 \\
        \midrule
        \multirow[t]{3}{*}{ \covid }
        & 5
        & \textbf{0.588} & \textbf{0.268}  
        & 0.659 & 0.314 &  
        \underline{0.640} & \underline{0.299}  
        & 1.241 & 0.691  
        & 1.117 & 0.646  
        & 1.259 & 0.675  
        \\
                & 10
        & \textbf{0.740} & \textbf{0.361}  
        & \underline{0.841} & \underline{0.410}  
        & 1.053 & 0.523  
        & 1.255 & 0.693  
        & 1.353 & 0.784  
        & 1.260 & 0.687  
        \\
                & 15
        & \textbf{0.932} & \textbf{0.461}  
        & \underline{1.026} & \underline{0.516}  
        & 1.309 & 0.686  
        & 1.265 & 0.690  
        & 1.351 & 0.792  
        & 1.277 & 0.718  
        \\
        \midrule
        \multirow[t]{3}{*}{\googletrend} & 5  
        & \textbf{0.573} & \textbf{0.442}  
        & \underline{0.626} & \underline{0.469}  
        & 0.719 & 0.551  
        & 1.255 & 1.024  
        & 0.919 & 0.640  
        & 1.038 & 0.981  
        \\
                & 10
        & \textbf{0.620} & \textbf{0.481}  
        & \underline{0.697} & \underline{0.514}  
        & 0.789 & 0.604  
        & 1.273 & 1.044  
        & 0.960 & 0.717  
        & 1.247 & 1.037  
        \\
                & 15
        & \textbf{0.646} & \textbf{0.505}  
        & \underline{0.701} & \underline{0.527}  
        & 0.742 & 0.571  
        & 1.300 & 1.069  
        & 0.828 & 0.631  
        & 1.038 & 0.795  
        \\
        \midrule 
        \multirow[t]{3}{*}{\chickendance} & 5
        & \textbf{0.353} & \textbf{0.221}  
        & 0.759 & 0.490  
        & \underline{0.492} & \underline{0.303}  
        & 0.890 & 0.767  
        & 0.508 & 0.316  
        & 2.037 & 1.742  
        \\
                & 10
        & \textbf{0.511} & \textbf{0.325}  
        & 0.843 & 0.564  
        & 0.838 & 0.535  
        & 0.886 & 0.753  
        & \underline{0.730} & \underline{0.476}  
        & 1.863 & 1.530  
        \\
                & 15
        & \textbf{0.653} & \textbf{0.419}  
        & 0.883 & 0.592  
        & 0.972 & 0.654  
        & \underline{0.862} & 0.718  
        & 0.903 & \underline{0.565}  
        & 1.792 & 1.481  
        \\
        \midrule
        \multirow[t]{3}{*}{\exercise} & 5
        & \textbf{0.309} & \textbf{0.177}  
        & 0.471 & \underline{0.275}  
        & 0.465 & 0.304  
        & \underline{0.408} & 0.290  
        & 0.424 & \underline{0.275}  
        & 1.003 & 0.748  
        \\
                & 10
        & \textbf{0.501} & \textbf{0.309}  
        & 0.630 & 0.381  
        & 0.789 & 0.518  
        & \underline{0.509} & 0.382  
        & 0.616 & \underline{0.377}  
        & 1.104 & 0.814  
        \\
                & 15
        & \underline{0.687} & \textbf{0.433}  
        & 0.786 & 0.505  
        & 1.147 & 0.758  
        & \textbf{0.676} & 0.475  
        & 0.691 & \underline{0.434}  
        & 1.126 & 0.901  
        \\
        \bottomrule
    \end{tabular}
    \label{table:forecasting_accuracy}
    \vspace{-1.1em}
\end{table*}
 
}
In this section, we evaluate the performance of \method using both synthetic and real-world datasets.
We performed extensive experiments to answer the following questions.
{\setlength{\leftmargini}{17.2pt}
\begin{enumerate}
    \renewcommand{\labelenumi}{\textit{Q\arabic{enumi}.}}
    \item 
    \textit{Effectiveness}: How well does it find the \relation?
    \item 
    \textit{Accuracy}: How accurately does it discover \relation and forecast future values?
    \item 
    \textit{Scalability}: How does it scale in terms of computational time?
\end{enumerate}
}
\myparaitemize{Datasets \& experimental setup}
We used the following datasets:
{\setlength{\leftmargini}{15pt}
\begin{itemize}
    \item
    \psynthetic:
    was generated based on a structural equation model~\cite{pearl2009causality}.
    Details
    are provided in Appendix \ref{section:app:experiments:setting}.
    \item
    \pcovid: was obtained from Google COVID-19 Open Data~\cite{googlecovid19}
    and consists of the number of infections in Japan, the United States, China, Italy, and the Republic of South Africa, covering over 900 daily entries.
    \item
    \pgoogletrend: consists of web-search counts collected over ten years related to beer queries on Google~\cite{gooogletrend}.
    \item
    \pchickendance, \pexercise: were obtained from the CMU motion capture database~\cite{cmumocap}
    and consist of four dimensional vectors (left/right legs and arms).
\end{itemize}
}
\noindent
We compared our algorithm with the following seven baselines for causal discovery, 
namely
CASPER~\cite{liu2023discovering},
DARING~\cite{he2021daring},
NoCurl~\cite{yu2021dag},
NOTEARS-MLP
(NO-MLP)
\cite{zheng2020learning},
NOTEARS~\cite{zheng2018dags},
LiNGAM~\cite{shimizu2006linear}, and
GES~\cite{chickering2002optimal}.
Besides, we also compared with the five following competitors for forecasting, namely
TimesNet~\cite{wu2023timesnet},
PatchTST~\cite{Yuqietal-2023-PatchTST},
DeepAR~\cite{salinas2020deepar},
OrbitMap~\cite{matsubara2019dynamic}, and
ARIMA~\cite{box1976arima}.
Details regarding the experimental
settings are also provided in Appendix \ref{section:app:experiments:setting}.
\par

\myparaitemize{Q1. Effectiveness}
We first demonstrated how effectively \method discovers the \relation and forecasts future values in a streaming fashion
using the epidemiological data stream (i.e., \covid).
Recall that Figure \ref{fig:crown} shows \method modeling and forecasting results.
Figure \ref{fig:crown} (a/b) shows graphical representations of the causal adjacency matrix $\mB$ and the eigenvalues $\eigs$.
Most importantly, the causal relationships evolve over time in accordance with
the transitions of distinct dynamical patterns in the inherent signals $\mE$.
\method can continuously detect new actual causative events around the world
(e.g., the discovery of a new lineage of the coronavirus in South Africa,
the abrupt increase in coronavirus infections in the United States,
and the strict, long-term lockdown in Shanghai).
Figure \ref{fig:crown} (c) shows streaming time series forecasting results. There have been multiple distinct patterns (e.g., a rapid decrease in infection numbers in the Republic of South Africa), \method adaptively captures the exponential patterns and forecasts future values close to the originals.
\par\noindent
\myparaitemize{Q2-1. Causal discovering accuracy}
We next showed how accurately \method can discover the \relation.
We reported the structural Hamming distance (SHD) and the structural intervention distance (SID)~\cite{peters2015structural}.
SHD quantifies the difference in the causal adjacency matrix by counting missing, extra, and reversed edges and
SID is particularly suited for evaluating causal discovering accuracy since it counts the number of couples $(i, j)$ such that the interventional distribution $p(x_j\mid\text{do}(X_i=\bar{x}))$ would be miscalculated if we used the estimated causal adjacency matrix.
Both metrics should be lower to represent better estimated adjacency matrices. 
Table \ref{table:discovering_accuracy} shows the causal discovering results of \method and its baselines for various synthetic datasets, where the best and second-best levels of performance are shown in \textbf{bold} and \underline{underlined}, respectively.
Our method outperformed all baselines for every temporal sequence, which is consistent with the analysis provided in Lemma \ref{lemma:causal}.
This is because none of the competitors can handle the \relation in data streams.
\par
\TSK{
\begin{table*}[t]
    \centering
    \caption{Ablation study results with forecasting steps $l_s\in\{5, 10, 15\}$ for both synthetic and real-world datasets.}
    \vspace{-1.0em}
    \begin{tabular}{c|c|cc|cc|cc|cc|cc}
    \toprule
    \multicolumn{2}{c|}{Datasets}
    & \multicolumn{2}{c|}{\synthetic} & \multicolumn{2}{c|}{\covid} & \multicolumn{2}{c|}{\googletrend} & \multicolumn{2}{c|}{\chickendance} & \multicolumn{2}{c}{\exercise} \\
    \midrule
    \multicolumn{2}{c|}{Metrics}
    & \:RMSE & MAE\:\,
    & \:RMSE & MAE\:\,
    & \:RMSE & MAE\:\,
    & \:RMSE & MAE\:
    & \:RMSE & MAE\:\: \\
    \midrule
    \multirow[t]{3}{*}{\:\:\method (full)\:\:}
    & 5 & \:0.722 & 0.528\:\, & \:0.588 & 0.268\:\, & \:0.573 & 0.442\:\, & \:0.353 & 0.221\: & \:0.309 & 0.177\:\, \\
    & 10 & \:0.829 & 0.607\:\, & \:0.740 & 0.361\:\, & \:0.620 & 0.481\:\, & \:0.511 & 0.325\: & \:0.501 & 0.309\:\, \\
    & 15 & \:0.923 & 0.686\:\, & \:0.932 & 0.461\:\, & \:0.646 & 0.505\:\, & \:0.653 & 0.419\: & \:0.687 & 0.433\:\, \\
    \midrule
    \multirow[t]{3}{*}{\:\:w/o causality\:\:}
    & 5 & \:0.759 & 0.563\:\, & \:0.758 & 0.374\:\, & \:0.575 & 0.437\:\, & \:0.391 & 0.262\: & \:0.375 & 0.218\:\, \\
    & 10 & \:0.925 & 0.696\:\, & \:0.848 & 0.466\:\, & \:0.666 & 0.511\:\, & \:0.590 & 0.398\: & \:0.707 & 0.433\:\, \\
    & 15 & \:1.001 & 0.760\:\, & \:1.144 & 0.583\:\, & \:0.708 & 0.545\:\, & \:0.821 & 0.537\: & \:0.856 & 0.533\:\, \\
    \bottomrule
    \end{tabular}
    \label{table:ablation}
    \vspace{-0.75em}
\end{table*}

}
\noindent
\myparaitemize{Q2-2. Forecasting accuracy}
We evaluated the quality of \method in terms of $l_s$-steps-ahead forecasting accuracy.
For this evaluation, we adopted the root mean square error (RMSE) and the mean absolute error (MAE),
both of which provide good results when they are close to zero.
For all methods, we used one-third of the sequences to tune their parameters.
Table \ref{table:forecasting_accuracy} presents the overall forecasting results,
where the best results are in \textbf{bold} and the second-best are \underline{underlined}.
For brevity, we only reported results of a representative synthetic dataset, which has the most complicated temporal sequence, ``1, 2, 3, 2, 1''.
We compared the two metrics
when we varied the forecasting step
(i.e., $l_s \in \{5, 10, 15\}$).
Our method achieved remarkable improvements over its competitors.
While deep learning models (TimesNet, PatchTST, and DeepAR)
exhibit high generality for time series modeling,
their forecasting accuracy was poorer
because they could not adjust the model parameters incrementally.
OrbitMap is capable of handling multiple discrete nonlinear dynamics but
misses the \relation,
and thus was outperformed by our proposed method.
ARIMA assumes linear relationships between time series data
and so fails to accommodate complex and nonlinear data
resulting in decreased forecasting accuracy.
\par
\myparaitemize{Q2-3. Ablation study}
To quantitatively evaluate the impact of causal relationships on forecasting effectiveness,
we additionally performed an ablation study by comparing a limited version of our method, namely \textit{w/o causality},
whose demixing matrix $\demixing$ was fixed to the identity matrix.
Table \ref{table:ablation} presents the overall results of our ablation study on \method using both synthetic and real-world datasets.
We can see that the \textit{w/o causality} causes a significant drop in forecasting accuracy across all experimental settings.
Therefore, we observed that the discovery of \relation in data streams boosts forecasting accuracy.
\TSK{
\begin{figure}[t]
    \centering
    \scalebox{1.01}{\includegraphics[width=\linewidth]{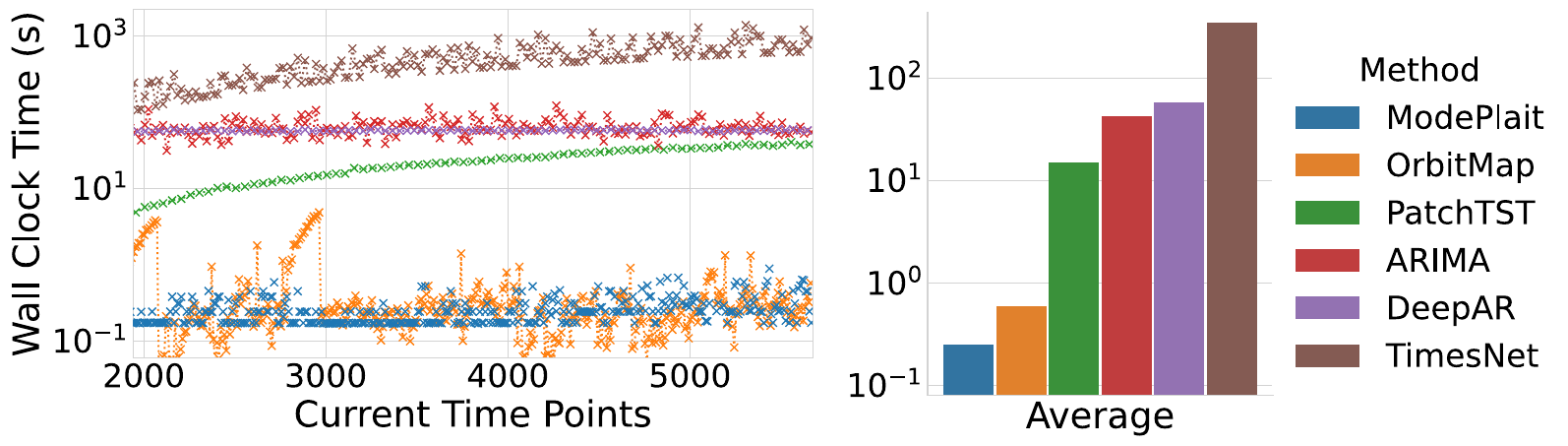}}
    \vspace{-5ex}
    \caption{Scalability of \method: (left) Wall clock time vs. data stream length $t_c$ and (right) average time consumption for \pexercise.
    The vertical axis of these graphs is a logarithmic scale.
    \method is superior to its competitors.
    It is up to 1,500x faster than its competitors.}
    \label{fig:time}
    \vspace{0.8ex}
\end{figure}

}
\par\noindent
\myparaitemize{Q3. Scalability}
Finally, we evaluated the computational time needed by our streaming algorithm.
Figure \ref{fig:time} compares the computational efficiencies
of \method and its competitors.
It presents the computational time at each time point $t_c$ on the left,
and the average computational time on the right.
Note that both figures are shown on linear-log scales.
Our method consistently outperformed its competitors
in terms of computational time
thanks to our incremental update,
which aligns with the discussion presented in Lemma \ref{lemma:stream_time}.
OrbitMap was competitive, but
it estimates model parameters via iterative optimization, the expectation-maximization algorithm,
which makes it slower than our proposed algorithm.
Other methods require a significant amount of learning time because
they cannot update their models incrementally.

\section{Conclusion}
    \label{section:conclusion}
    In this paper,
we focused on the summarization of an entire data stream, discovering the \relation in data streams, and
forecasting future values incrementally.
Our proposed method, namely \method, exhibits all of the following desirable properties that we listed in the introduction:
{\setlength{\leftmargini}{15pt}
\begin{itemize}
    \item It is \textit{Effective}: It provides the \relation, namely insightful time-changing causal relationships in data streams.
    \item It is \textit{Accurate}: Our experiments demonstrated that \method precisely discovers the \relation and forecasts future values in a streaming fashion.
    \item It is \textit{Scalable}: The computational time for our proposed algorithm does not depend on the data stream length.
\end{itemize}}

\myparaitemize{Acknowledgment}
We would like to thank the anonymous referees
for their valuable and helpful comments.
This work was partly supported by
``Program for Leading Graduate Schools'' of the Osaka University, Japan, 
JST CREST JPMJCR23M3,
JSPS KAKENHI Grant-in-Aid for Scientific Research Number
JP24KJ1618.    


\bibliographystyle{bib/ACM-Reference-Format}
\bibliography{bib/references}

\clearpage
\appendix
\section*{Appendix}
\label{appendix}

\section{Optimization Algorithm}
\TSK{
\begin{figure}[!h]
\vspace{-5.0ex}
\begin{algorithm}[H]
    \normalsize
    \caption{\method($\vx(t_c), \modelparam, \candparam$)}
    \label{alg:model}
    \begin{algorithmic}[1]
        \STATE {\bf Input:}
        \hspace{0mm}    (a) New value $\vx(t_c)$ at time point $t_c$ \\
        \hspace{9.5mm} (b) Full parameter set $\modelparam = \{\regimeset, \updateset\}$ \\
        \hspace{9.68mm} (c) Model candidate $\candparam = \{\regime^c, \update^c, \bm{s}^c_{en}\}$
        \STATE {\bf Output:}
        \hspace{0mm}    (a) Updated full parameter set $\modelparam'$ \\
        \hspace{11.8mm} (b) Updated model candidate $\candparam'$ \\
        \hspace{11.9mm} (c) $l_s$-steps-ahead future value $\vect{v}(t_c+l_s)$ \\
        \hspace{11.8mm} (d) Causal adjacency matrix $\mB$
        \STATE /* Update current window $\mX^c$ */
        \STATE $\mX^c \leftarrow \mX[t_m : t_c]$
        \STATE /* Estimate optimal regime $\regime$ */
        \STATE $\{\modelparam', \candparam'\} \leftarrow$ \modelestimator($\mX^c, \modelparam$, $\candparam$)
        \STATE /* Forecast future value and discover causal relationship */
        \STATE $\{\vect{v}(t_c+l_s),~\mB\} \leftarrow$ \modelgenerator($\candparam'$)
        \STATE /* Update regime $\regime$ */
        \IF{NOT create new regime}
            \STATE $\candparam' \leftarrow \regimeupdate(\mX^c, \candparam')$
        \ENDIF
    \RETURN $\{\modelparam', \candparam', \vect{v}(t_c+l_s), \mB\}$
    \end{algorithmic}
\end{algorithm}
\vspace{-4.5em}
\end{figure}
}\par
\TSK{
\begin{figure}[!h]
\vspace{-0.0ex}
\begin{algorithm}[H]
    \normalsize
    \caption{\modelestimator($\mX^c, \modelparam, \candparam$)}
    \label{alg:estimator}
    \begin{algorithmic}[1]
        \STATE {\bf Input:}
        \hspace{0.0mm}  (a) Current window $\mX^c$ \\
        \hspace{9.5mm} (b) Full parameter set $\modelparam$ \\
        \hspace{9.68mm} (c) Model candidate $\candparam$
        \STATE {\bf Output:}
        \hspace{0.0mm}  (a) Updated full parameter set $\modelparam'$ \\
        \hspace{11.8mm} (b) Updated model candidate $\candparam'$
        \STATE /* Calculate optimal initial conditions */
        \STATE $\mat{S}_{0}^c \leftarrow \argmin_{\mat{S}_{0}^c} f(\mX^c; \mat{S}_{0}^c, \regime^c)$
        \IF{$f(\mX^c; \mat{S}_{0}^c, \regime^c) > \tau$}
            \STATE /* Find better regime in $\bm{\Theta}$ */
            \STATE $\{ \mat{S}_{0}^c, \regime^c \} \leftarrow \argmin_{\mat{S}_{0}^c, \regime \in \regimeset} \,f(\mX^c; \mat{S}_{0}^c, \regime^c)$
            \IF{$f(\mX^c; \mat{S}_{0}^c, \regime^c) > \tau$}
                \STATE /* Create new regime */
                \STATE $\{ \regime^c, \update^c \} \leftarrow \textsc{RegimeCreation}(\mX^c)$
                \STATE $\regimeset \leftarrow \regimeset \cup \regime^c$; $\updateset \leftarrow \updateset \cup \update^c$
            \ENDIF
        \ENDIF
        \STATE $\modelparam' \leftarrow \{\regimeset, \updateset\}$; $\candparam' \leftarrow \{\regime^c, \update^c, \mat{S}_{en}^c\}$
        \RETURN $\modelparam', \candparam'$
    \end{algorithmic}
\end{algorithm}
\vspace{-3.3em}
\end{figure}
}
\label{section:app:algorithm}
\subsection{Details of Eq. (\ref{eq:update_trans})}
Here, we introduce the recurrence relation of transition matrix $\ith{\trans}$.
As mentioned earlier, we use the following cost function (below, index $i$ denoting $i$-th dimension is omitted for the sake of simplicity, e.g., we write $\ith{\trans}$ as $\trans$):
\begin{align*}
    \mathcal{E} &= \sum_{t'=t_m+h}^{t_c}\forgetting^{t_c-t'}||\embed{e(t')} - \trans\embed{e(t'-1)}||_2^2 \\
    &= \sum_{l=1}^h (\mat{L}(l, :) - \trans(l, :)\mat{R})\Forgetting(\mat{L}(l, :) - \trans(l, :)\mat{R})^\top
\end{align*}
where,
$\Forgetting, \mat{L}$ and $\mat{R}$ are synonymous with the definition in Section \ref{section:alg:creation}.
Because we want to obtain $\trans$ that minimizes this cost function $\mathcal{E}$, we differentiate it with respect to $\trans$.
\begin{align*}
    \dfrac{\partial}{\partial\trans(l, :)}\mathcal{E}
    &= -2(\mat{L}(l, :) - \trans(l, :)\mat{R}) \Forgetting \mat{R}^\top
\end{align*}
Solving the equation $\partial\mathcal{E}/\partial\trans(l, :) = 0$ for each $l$, $1 \leq l \leq h$,
the optimal solution for $\trans$ is given by
$$ \trans = (\mat{L}\Forgetting\mat{R}^\top)(\mat{R}\Forgetting\mat{R}^\top)^{-1} $$
where we define
\begin{align*}
    \mat{Q} = \mat{L}\Forgetting\mat{R}^\top,\quad
    \mat{P} = (\mat{R}\Forgetting\mat{R}^\top)^{-1}
\end{align*}
The recurrence relations of $\mat{Q}$ can be written as
\begin{align*}
    \mat{Q} &= \sum_{t'=t_m+h}^{t_c}\forgetting^{t_c-t'}\embed{e(t')}\embed{e(t'-1)}^\top \\
    &= \forgetting\sum_{t'=t_m+h}^{t_c-1}\forgetting^{t_c-t'-1}\embed{e(t')}\embed{e(t'-1)}^\top + \embed{e(t_c)}\embed{e(t_c-1)}^\top
\end{align*}
\begin{align}
    \label{eq:Q}
    \therefore \mat{Q}^{new} = \forgetting\mat{Q}^{prev} + \embed{e(t_c-1)}\embed{e(t_c)}^\top
\end{align}
and similarly
\begin{align}
    \label{eq:bP}
    \mat{P}^{new} &= (\forgetting{(\mat{P}^{prev})}^{-1} + \embed{e(t_c)}\embed{e(t_c)}^\top)^{-1}
\end{align}Here, we apply the Sherman-Morrison formula~\cite{sherman1950adjustment} to the RHS of Eq. \eqref{eq:bP}.
Note that $\embed{e(t_c)}^\top\mat{P}^{prev}\embed{e(t_c)} > 0$
because $\mat{P}^{-1} = \mat{R}\Forgetting\mat{R}^\top$ is positive definite by definition.
\begin{align}
    \label{eq:P}
    \therefore \mat{P}^{new} = \frac{1}{\forgetting}(\mat{P}^{prev} - \frac{\mat{P}^{prev}\embed{e(t_c-1)}\embed{e(t_c-1)}^\top\mat{P}^{prev}}{\forgetting + \embed{e(t_c-1)}^\top\mat{P}^{prev}\embed{e(t_c-1)}})
\end{align}
Finally, combining Eq. \eqref{eq:Q} and Eq. \eqref{eq:P} gives the recurrence relations of $\trans$ for Eq. \eqref{eq:update_trans}.
\begin{align*}
    \begin{split}
        \trans^{new} &= \trans^{prev} + (\embed{e(t_c)} - \trans^{prev}\embed{e(t_c-1)}\boldsymbol\gamma \\
        \boldsymbol\gamma &= \frac{\embed{e(t_c-1)}^\top\mat{P}^{prev}}{\forgetting + \embed{e(t_c-1)}^\top\mat{P}^{prev}\embed{e(t_c-1)}}
    \end{split}
\end{align*}
    
\par
\setcounter{lemma}{1}
\subsection{Proof of Lemma \ref{lemma:create_time}}
\begin{proof}
The dominant steps in \textsc{RegimeCreation} are I, IV, and VI.
The decomposition $\mX$ into $\demixing^{-1}$ and $\mE$ using ICA requires $O(d^2N)$.
For each observation,
the SVD of $\ith{\mat{R}}\mat{M}$ requires $O(h^2N)$, and the eigendecomposition of $\ith{\tilde{\trans}}$ takes $O(k_i^3)$.
The straightforward way to
process IV and VI
is to perform the calculation $d$ times sequentially, i.e., they require $O(dh^2N+\sum_ik_i^3)$ in total.
However, since these operations do not interfere with each other,
they are simultaneously computed by parallel processing.
Therefore, the time complexity of \textsc{RegimeCreation} is $O(N(d^2+h^2)+k^3)$, where $k=\max_i(k_i)$.
\end{proof}
\subsection{Proof of Lemma \ref{lemma:causal}}
\begin{proof}
First, we need to formulate the causal structure.
Here, we utilize the structural equation model~\cite{pearl2009causality}, denoted by $\mX_{\text{sem}} = \mB_{\text{sem}}\mX_{\text{sem}} + \mE_{\text{sem}}$.
Because this model is known as the general formulation of causality, if $\mB_{\text{sem}}$ in this model is identified, then it can be said that we discover causality.
In other words, we need to prove that our proposed algorithm can find the causal adjacency matrix $\mB$ aligning with this model.
Solving the structural equation model for $\mX_{\text{sem}}$, we obtain 
$\mX_{\text{sem}} = \demixing^{-1}_{\text{sem}}\mE_{\text{sem}}$
where $\demixing_{\text{sem}} = \mat{I} - \mB_{\text{sem}}$.
It is shown that we can identify $\demixing_{\text{sem}}$ in the above equation by ICA,
except for the order and scaling of the independent components, if the observed data is a linear, invertible mixture of non-Gaussian independent components~\cite{comon1994independent}.
Thus, demonstrating that \modelgenerator precisely resolves the two indeterminacies of a mixing matrix $\mW^{-1}$ (i.e., the inverse of $\demixing \in \regime^c$) suffices to complete the proof because $\demixing$ is computed by ICA in \textsc{RegimeCreation}. \par
First, we reveal that our algorithm can resolve the order indeterminacy.
We can permutate the causal adjacency matrix $\mB$ to strict lower triangularity thanks to the acyclicity assumption~\cite{bollen1989structural}.
Thus, correctly permuted and scaled $\mW$
is a lower triangular matrix with all ones on the diagonal.
It is also said that there would only be one way to permutate $\mW$, which meets the above condition~\cite{shimizu2006linear}.
Thus, \modelgenerator can identify the order of a mixing matrix by the process in step I (i.e., finding the permutation of rows of a mixing matrix that yields a matrix without any zeros on the main diagonal).
Next, with regard to the scale of indeterminacy,
it is apparent that we only need to focus on the diagonal element,
remembering that the permuted and scaled $\mW$ has all ones on the diagonal.
Therefore, we prove that \modelgenerator can resolve the order and scaling of the indeterminacies of a mixing matrix $\demixing^{-1}$.
\end{proof}
\subsection{Proof of Lemma \ref{lemma:stream_time}}
\begin{proof}
For each time point, \method first runs \modelestimator,
which estimates the optimal full parameter set $\modelparam$ and the model candidate $\candparam$ for the current window $\mX^c$.
If the current regime $\regime^c$ fits well,
it takes $O(N\sum_i k_i)$ time.
Otherwise, it takes $O(RN\sum_i k_i)$ time to find a better regime in $\regimeset\in\modelparam$.
Furthermore, if \method encounters an unknown pattern,
it runs \textsc{RegimeCreation}, which takes $O(N(d^2+h^2)+k^3)$ time.
Subsequently, it runs \modelgenerator to identify the causal adjacency matrix and forecast an $l_s$-steps-ahead future value,
which takes $O(d^2)$ and $O(l_s)$ time, respectively.
Note that $l_s$ is negligible because of the small constant value.
Finally, when \method does not create a new regime,
it executes \regimeupdate, which needs $O(dh^2)$ time.
Thus, the total time complexity is at least $O(N\sum_ik_i+dh^2)$ time and at most $O(RN\sum_i k_i+N(d^2+h^2)+k^3)$ time per process.
\end{proof}

\section{Experimental Setup}
\label{section:app:experiments:setting}
In this section, we describe the experimental setting in detail.
We conducted all our experiments on
an Intel Xeon Platinum 8268 2.9GHz quad core CPU
with 512GB of memory and running Linux.
We normalized the values of each dataset based on their mean and variance (z-normalization).
The length of the current window $N$ was $50$ steps in all experiments.
\par
\myparaitemize{Generating the Datasets}
We randomly generated synthetic multivariate data streams containing multiple clusters, each of which exhibited a certain causal relationship.
For each cluster, the causal adjacency matrix $\mB$ was generated from a well-known random graph model, namely Erdös-Rényi (ER)~\cite{erdos1960evolution} with edge density $0.5$ and the number of observed variables $d$ was set at 5.
The data generation process was modeled as a structural equation model~\cite{pearl2009causality},
where each value of the causal adjacency matrix $\mB$ was sampled from a uniform distribution $\mathcal{U}(-2, -0.5)\cup(0.5, 2)$.
In addition, to demonstrate the time-changing nature of exogenous variables, 
we allowed the inherent signals variance $\sigma^2_{i, t}$ (i.e., $\ith{e}(t)\sim\text{Laplace}(0, \sigma_{i, t}^2)$)
to change over time.
Specifically, we introduced $h_{i, t}=\text{log}(\sigma^2_{i, t})$, which evolves according to an autoregressive model, where the coefficient and noise variance of the autoregressive model were sampled from $\mathcal{U}(0.8, 0.998)$ and $\mathcal{U}(0.01, 0.1)$, respectively.

The overall data stream was then generated by constructing a temporal sequence of cluster segments and each segment had $500$ observations (e.g., ``$1,2,1$'' consists of three segments containing two types of causal relationships and its total sample size is $1,500$). We ran our experiments on five different temporal sequences: ``$1,2,1$'', ``$1,2,3$'', ``$1,2,2,1$'', ``$1,2,3,4$'', and ``$1,2,3,2,1$'' to encompass various types of real-world scenarios.
\par
\myparaitemize{Baselines}
The details of the baselines we used throughout our extensive experiments are summarized as follows:
\par\noindent
(1) Causal discovering methods
{\setlength{\leftmargini}{11pt}
\vspace{-0.3ex}
\begin{itemize}
    \item CASPER~\cite{liu2023discovering}: is a state-of-the-art method for causal discovery, integrating the graph structure into the score function and reflecting the causal distance between estimated and ground truth causal structure. We tuned the parameters by following the original paper setting.
    \item DARING~\cite{he2021daring}: introduces an adversarial learning strategy to impose an explicit residual independence constraint for causal discovery. We searched for three types of regularization penalties $\{\alpha, \beta, \gamma\}\in\{0.001, 0.01, 0.1, 1.0, 10\}$.
    \item NoCurl~\cite{yu2021dag}: uses a two-step procedure: initialize a cyclic solution first and then employ the Hodge decomposition of graphs. We set the optimal parameter presented in the original paper.
    \item NOTEARS-MLP~\cite{zheng2020learning}: is an extension of NOTEARS~\cite{zheng2018dags} (mentioned below) for nonlinear settings, which aims to approximate the generative structural equation model by MLP.
    We used the default parameters provided in authors' codes\footnote[2]{\url{https://github.com/xunzheng/notears} \label{fot:notears}}.
    \item NOTEARS~\cite{zheng2018dags}:
    is a differentiable optimization method with an acyclic regularization term to estimate a causal adjacency matrix.
    We used the default parameters provided in authors' codes\footref{fot:notears}.
    \item LiNGAM~\cite{shimizu2006linear}:
    exploits the non-Gaussianity of data to determine the direction of causal relationships. It has no parameters to set.
    \item GES~\cite{chickering2002optimal}: is a traditional score-based bayesian algorithm that discovers causal relationships in a greedy manner.
    It has no parameters to set.
    We employed BIC as the score function and utilized the open-source in~\cite{kalainathan2020causal}.
\end{itemize}
\vspace{-0.5ex}}
\par\noindent
(2) Time series forecasting methods
{\setlength{\leftmargini}{11pt}
\vspace{-0.3ex}
\begin{itemize}
    \item TimesNet/PatchTST~\cite{wu2023timesnet, Yuqietal-2023-PatchTST}: are state-of-the-art TCN-based and Transformer-based methods, respectively.
    The past sequence length was set as 16 (to match the current window length).
    Other parameters followed the original paper setting.
    \item DeepAR~\cite{salinas2020deepar}: models probabilistic distribution in the future, based on RNN. We built the model with 2-layer 64-unit RNNs. We used Adam optimization~\cite{adam} with a learning rate of 0.01 and let it learn for 20 epochs with early stopping.
    \item OrbitMap~\cite{matsubara2019dynamic}:
    finds important time-evolving patterns for stream forecasting.
    We determined the optimal transition strength $\rho$ to minimize the forecasting error in training.
    \item ARIMA~\cite{box1976arima}: is one of the traditional time series forecasting approaches based on linear
    equations. We determined the optimal parameter set using AIC.
\end{itemize}}

\end{document}